\documentclass[letterpaper]{article} 
\usepackage[draft]{aaai2026}  
\usepackage{times}  
\usepackage{helvet}  
\usepackage{courier}  
\usepackage[hyphens]{url}  
\usepackage{graphicx} 
\usepackage{natbib}  
\usepackage{caption} 
\usepackage{algorithm}
\usepackage{algorithmic}
\usepackage{amsmath}
\usepackage{amssymb}
\usepackage{booktabs}
\usepackage{multirow}
\usepackage{graphicx}
\usepackage{newfloat}
\usepackage{listings}
\DeclareCaptionStyle{ruled}{labelfont=normalfont,labelsep=colon,strut=off} 
\floatstyle{ruled}
\newfloat{listing}{tb}{lst}{}
\floatname{listing}{Listing}
\title{Learning Faithful Reasoning Processes with Spatio-Temporal Latent Dynamics for Large Language Models}

\author{
    Weijiang Lv\textsuperscript{\rm 1}\equalcontrib,
    Wentong Zhao\textsuperscript{\rm 1}\equalcontrib,
    Jiayu Wang\textsuperscript{\rm 2},
    Yuhao Wu\textsuperscript{\rm 3},
    Jiaheng Wei\textsuperscript{\rm 4},
    Xiaobo Xia\textsuperscript{\rm 5}\thanks{Corresponding author (xiaoboxia@ustc.edu.cn).}
}
\affiliations{
    \textsuperscript{\rm 1}Xidian University\\
    \textsuperscript{\rm 2}Xi'an Jiaotong University\\
    \textsuperscript{\rm 3}Mohamed bin Zayed University of Artificial Intelligence\\
    \textsuperscript{\rm 4}The Hong Kong University of Science and Technology (Guangzhou)\\
    \textsuperscript{\rm 5}University of Science and Technology of China
}

\usepackage{bibentry}

\begin{document}

\maketitle

\begin{abstract}
Chain-of-Thought (CoT) reasoning has significantly enhanced the reasoning capabilities of large language models (LLMs). However, existing reinforcement learning methods primarily optimize outcome-level rewards, providing limited supervision over the reasoning process. As a result, models may produce shortcut-prone or unstable reasoning trajectories even when the final answer is correct. We propose GeoFaith, a spatio-temporal latent dynamics framework for learning faithful reasoning processes without process reward models. GeoFaith combines final-state latent uncertainty, which measures whether a completed trajectory lies in a well-supported manifold region, with predictive-entropy dynamics, which penalize abnormal uncertainty patterns such as premature collapse, oscillation, and prolonged uncertainty. These signals are incorporated as process regularization into GRPO, jointly optimizing answer correctness and reasoning quality without step-level process annotations. Experiments  show that GeoFaith consistently improves reasoning accuracy, produces more efficient reasoning trajectories, and outperforms both outcome-only reinforcement learning and process-supervision baselines.

\end{abstract}


\section{Introduction}

Chain-of-Thought (CoT) reasoning~\cite{wei2022chbain} has significantly improved the ability of large language models (LLMs) to solve complex reasoning tasks by eliciting intermediate reasoning steps~\cite{chen2025towards,zhao2025cot,zawalski2024robotic,feng2023towards}. Recent reinforcement learning methods further enhance reasoning performance by optimizing final-answer correctness~\cite{shao2024deepseekmath,wen2025policy}. However, outcome-level rewards provide only sparse supervision: they indicate whether the final answer is correct, but offer limited guidance on how the reasoning process should unfold. As a result, models may arrive at correct answers through shortcut-prone, unstable, or poorly grounded reasoning trajectories.

A natural solution is to introduce process-level supervision, such as human annotations, LLM-as-judge feedback, or process reward models (PRMs). These approaches can improve reasoning quality, but they often require expensive step-level labels, inherit judge bias, or introduce additional learned verifiers whose reliability may not generalize across domains. More importantly, process supervision is usually imposed from the outside, while the model's own internal reasoning dynamics remain underutilized. This raises a central question: \emph{can we improve reasoning processes without training a separate process reward model or collecting step-level process annotations?}

In this work, we approach reasoning process optimization through the lens of \textit{spatio-temporal latent dynamics}. Instead of treating CoT as only a sequence of surface-level tokens, we analyze how reasoning trajectories evolve in the model's hidden-state space and how predictive uncertainty changes across reasoning steps. Reliable reasoning trajectories tend to exhibit coherent latent geometry and stable uncertainty reduction, whereas unreliable trajectories often show geometric deviations, premature entropy collapse, oscillation, or prolonged uncertainty. These observations motivate a view of process supervision: reasoning quality can be regularized directly through internal trajectory signals.

We propose \textbf{GeoFaith}, a spatio-temporal latent dynamics framework for reasoning process verification without process reward models. GeoFaith uses two complementary process rewards. First, a \textit{spatial latent-geometry} reward favors completed trajectories whose final hidden state lies in a well-supported manifold region. Second, a \textit{temporal predictive-entropy} reward penalizes abnormal uncertainty patterns such as premature collapse, oscillation, and prolonged uncertainty. We incorporate these signals into GRPO, jointly optimizing final-answer correctness and reasoning-process quality without step-level process annotations.

The main contributions of this work are summarized as follows:

\begin{itemize}
    \item We propose a spatio-temporal latent dynamics perspective for reasoning process optimization, modeling CoT trajectories through hidden-state geometry and predictive-entropy evolution.
    \item We introduce detector-free process regularization signals based on manifold consistency and entropy dynamics, avoiding step-level annotations and learned process reward models.
    \item We integrate these signals into GRPO to jointly optimize answer correctness and reasoning quality, improving accuracy, efficiency, and process reliability over outcome-only and other RL baselines.
\end{itemize}

\begin{figure}[t]
  \centering
  \includegraphics[width=\linewidth]{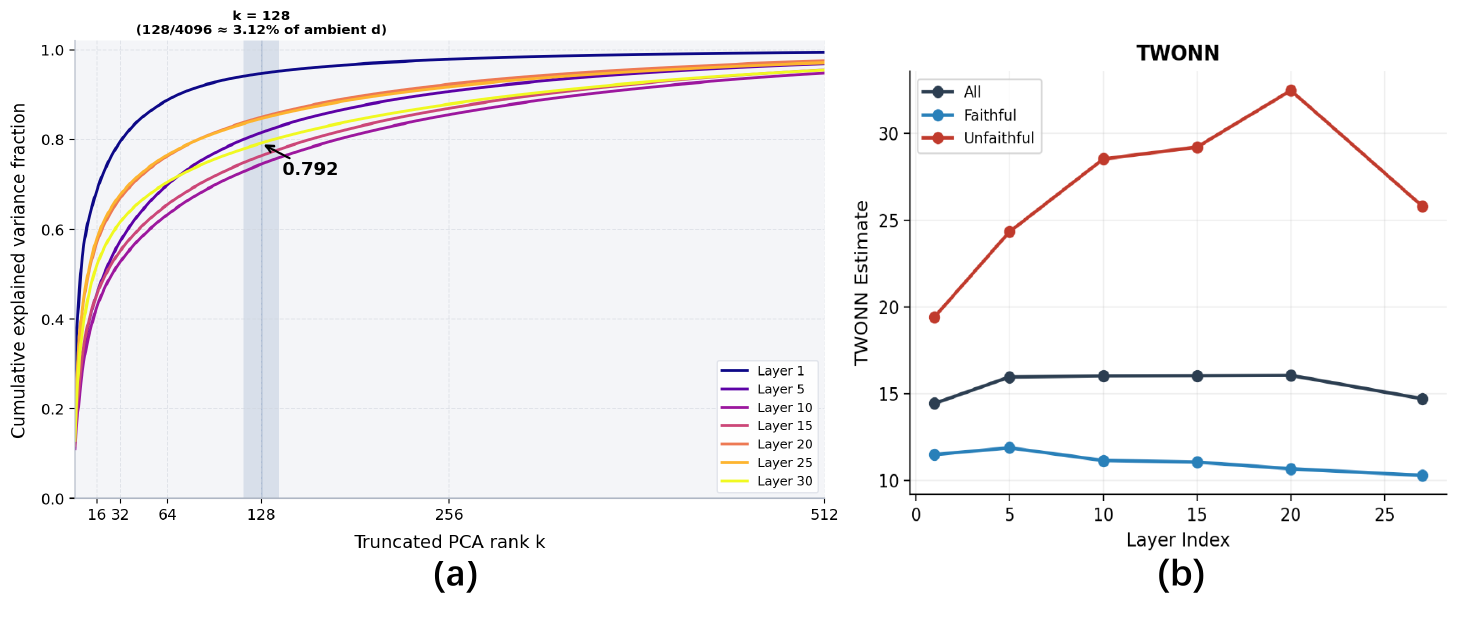}
  \caption{\textbf{High-dimensional CoT hidden states concentrate on a low-dimensional structured manifold.} \textit{Left}: layer-wise PCA shows that a small number of components explain most variance; \textit{Right}: TwoNN intrinsic-dimension estimates remain far below the ambient dimension across layers.}
  \label{fig:pca-twonn}
\end{figure}%

\section{Related Work}


\subsection{Faithfulness in CoT Reasoning}
Recent work has shown that CoT reasoning is not always faithful to the model's actual decision process~\cite{turpin2023language,lanham2023measuring,lv2026spd,xu2024faithful}. We distinguish \emph{causal faithfulness}, where the visible CoT reflects the internal computations that caused the answer, from \emph{observable faithfulness}, where the generated reasoning is logically valid, evidence-grounded, and answer-consistent. GeoFaith targets this observable notion of process quality, which is the level at which CoT traces can be evaluated and optimized in our setting. Existing methods include consistency-based~\cite{wang2022self}, intervention-based~\cite{lanham2023measuring}, and judge-based approaches. They face shortcut repetition~\cite{mehta2026does}, high intervention cost, judge calibration issues, and reward hacking when used as RL rewards. GeoFaith instead uses latent geometry and predictive-entropy dynamics as model-internal proxies for observable process quality.

\subsection{Representations of Reasoning Trajectories}
Prior studies have shown that intermediate reasoning states form step-specific trajectories that become increasingly structured at deeper layers~\cite{valeriani2023geometry,sun2026llm,ma2026reasoning}. Besides, successful reasoning is encoded in trajectory geometry, e.g., speed and curvature, rather than final representations alone~\cite{marin2024geometric,jiang2026beyond}. Concurrent work has examined reasoning faithfulness, demonstrating that models often overwrite intermediate representations or perform hidden reasoning beneath superficial tokens~\cite{turpin2023language,lanham2023measuring}. To steer models toward faithful regions, GeoSteer learns a low-dimensional manifold of high-quality reasoning trajectories~\citep{kazama2026geosteer}.

\begin{figure}[t]
  \centering
  \includegraphics[width=\linewidth]{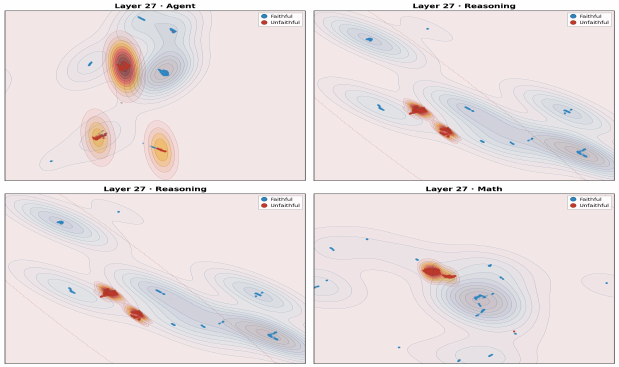}
  \caption{\textbf{Geometric separability of reasoning quality.} UMAP visualizations of hidden states across layers (columns) and task domains (rows). Blue points and density contours denote faithful reasoning; red points and contours denote unfaithful reasoning.}
  \label{fig:umap}
\end{figure}

\section{Spatio-Temporal Latent Dynamics of Reasoning}
\subsection{Preliminaries}
Given a query $x \in \mathcal X$, a large language model (LLM) $\pi_{\theta}$ generates a reasoning
chain $\tau=(c_{1},\ldots,c_{T})$, where $c_t$ denotes the $t$-th reasoning
step. Let $\mathbf h_t^\ell \in \mathbb R^{d_h}$ denote the last-token hidden
state at layer $\ell$ after generating $c_t$. The sequence
$\mathbf H_\tau^\ell=(\mathbf h_1^\ell,\ldots,\mathbf h_T^\ell)$ defines the corresponding latent trajectory.

\subsection{Spatial Geometry of Reasoning Processes}
\label{sec:spatial-properties}
\textbf{Hidden states are low-dimensional.}
We aggregate hidden states across reasoning
trajectories and perform layer-wise geometric
analysis. As shown in Figure~\ref{fig:pca-twonn},
the cumulative explained variance saturates
rapidly: fewer than 128 principal components
($\approx 3\%$ of the ambient dimension $d_h$)
explain most of the variance.  We further estimate
intrinsic dimensionality using
TwoNN~\citep{ansuini2019intrinsic}, obtaining
consistently low values across layers. Together,
these results suggest that hidden states
concentrate near a low-dimensional manifold
within the hidden-state space. However, the ambient space \(\mathbb{R}^{d_h}\) remains strongly anisotropic, so Euclidean distances and linear separators there can reflect layer-specific scale rather than reasoning quality. Following GeoSteer~\citep{kazama2026geosteer}, we map hidden states into a VAE latent chart before geometric analysis.

\noindent\textbf{Topological Divergence of Reasoning Quality.}
Based on the UMAP visualizations~\citep{mcinnes2018umap}
in Figure~\ref{fig:umap}, reasoning trajectories with different
process quality exhibit clear topological divergence in the latent
space across mathematical, logical, and knowledge-intensive domains. More visualizations can be found in Appendix A.2 . This suggests that hidden-state neighborhood structures are associated with reasoning quality.

To construct the low-quality contrast set for visualization, we use
RFEval-derived counterfactual reasoning traces~\citep{han2026rfeval}.
Each RFEval instance provides a problem, a plausible but flawed
counterfactual reasoning trace, and an explanation of the injected
reasoning error. We treat these counterfactual traces as low-quality
trajectories and compare them with trajectories that preserve
reasoning-answer consistency. Additional visualizations are provided in the supplementary material.

\begin{figure}[t]
  \centering
  \includegraphics[width=\linewidth]{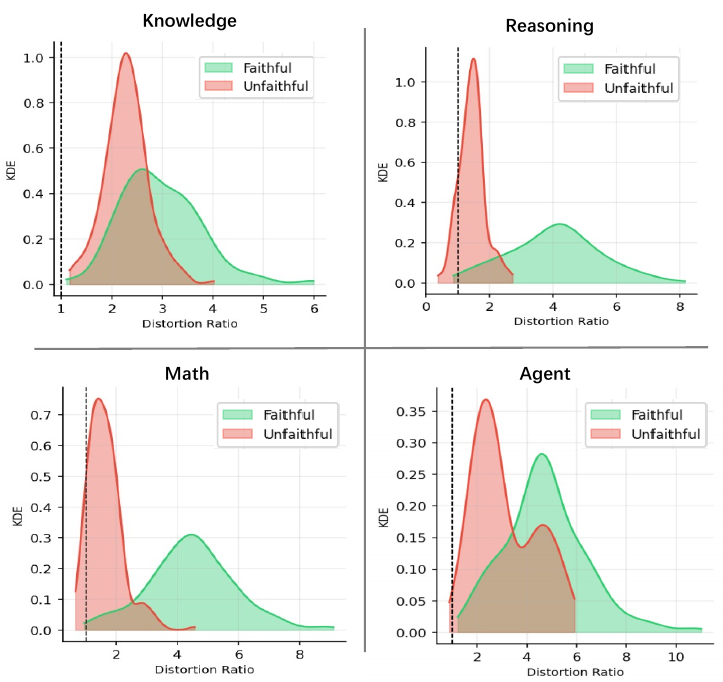}
  \caption{Distortion ratio distributions for faithful, unfaithful, and cross-class pairs.}
  \label{fig:distortion}
\end{figure}

To further quantify this geometric separation, we leverage the Riemannian geometry of the Variational Autoencoder~(VAE) latent space~\citep{burgess2018understanding}. Let $f:\mathbb R^{d_z}\rightarrow\mathbb R^{2d_h}$ denote the decoder mapping $\mathbf{z} \mapsto [
\boldsymbol\mu(\mathbf z);
\boldsymbol\sigma(\mathbf z)
]$. We define a well-established method of pullback metric on the latent space induced by the VAE decoder~\citep{arvanitidis2017latent,syrota2024decoder}
\begin{equation}
\mathbf G(\mathbf z)
=
\frac{1}{M}
\sum_{a=1}^{M}
\mathbf J_f^{(a)}(\mathbf z)^\top
\mathbf J_f^{(a)}(\mathbf z),
\end{equation} where $\mathbf J_f^{(a)}(\mathbf z)$ denotes the Jacobian of the $a$-th decoder. Under this metric, the length of a smooth path $\gamma: [0, 1] \to \mathbb{R}^{d_z}$ is 
$
L(\gamma)
=
\int_0^1
\sqrt{
\dot\gamma(t)^\top
\mathbf G(\gamma(t))
\dot\gamma(t)
}
\,dt,
$ where $\dot\gamma(t)$ denotes the local direction of movement along the trajectory at time $t$.
The corresponding Riemannian distance between
latent states $\mathbf z_i$ and $\mathbf z_j$ is defined as the infimum of $L(\gamma)$ over all paths connecting them.

In practice, we approximate the continuous path length using a $k$-nearest-neighbor graph constructed on latent representations. For each edge $(i,j)$, define
$
\boldsymbol\Delta_{ij}
=
\mathbf z_j-\mathbf z_i,
\quad
\bar{\mathbf z}_{ij}
=
\frac{\mathbf z_i+\mathbf z_j}{2}.
$
By discretizing the Riemannian line element
$
ds^2 = d\mathbf z^\top \mathbf G(\mathbf z)\, d\mathbf z,
$
where
$
d\mathbf z = \dot\gamma(t)\,dt,
$
the corresponding edge weight is
\begin{equation}
w_{ij}
=
\sqrt{
\max\!\left(
0,
\boldsymbol\Delta_{ij}^{\top}
\mathbf G(\bar{\mathbf z}_{ij})
\boldsymbol\Delta_{ij}
\right)
+\varepsilon},
\end{equation}
where $\varepsilon$ is a small numerical constant for stability. The discrete geodesic distance is then defined as
$
d_{\mathrm{geo}}(\mathbf z_i,\mathbf z_j)
=
\min_{p\in\mathcal P_{ij}}
\sum_{(u,v)\in p}
w_{uv}.
$
We further define the Euclidean distance as $d_{\mathrm{euc}}(\mathbf z_i,\mathbf z_j)
=
\|\mathbf z_i-\mathbf z_j\|_2$ and the distortion ratio as $\rho_{ij}
=
\frac{
d_{\mathrm{geo}}(\mathbf z_i,\mathbf z_j)
}{
d_{\mathrm{euc}}(\mathbf z_i,\mathbf z_j)
}.$ Figure~\ref{fig:distortion} illustrates that low-quality trajectories can exhibit lower distortion ratios \(\rho_{ij}\), indicating more direct Euclidean paths through the latent manifold. We use this distribution-level analysis to motivate the spatial signal; the per-sample score in Section~\ref{process-verification} uses final-state uncertainty rather than distortion. More VAE training details are provided in Appendix A.3.

\begin{figure*}[t]
  \centering
  \includegraphics[width=\textwidth]{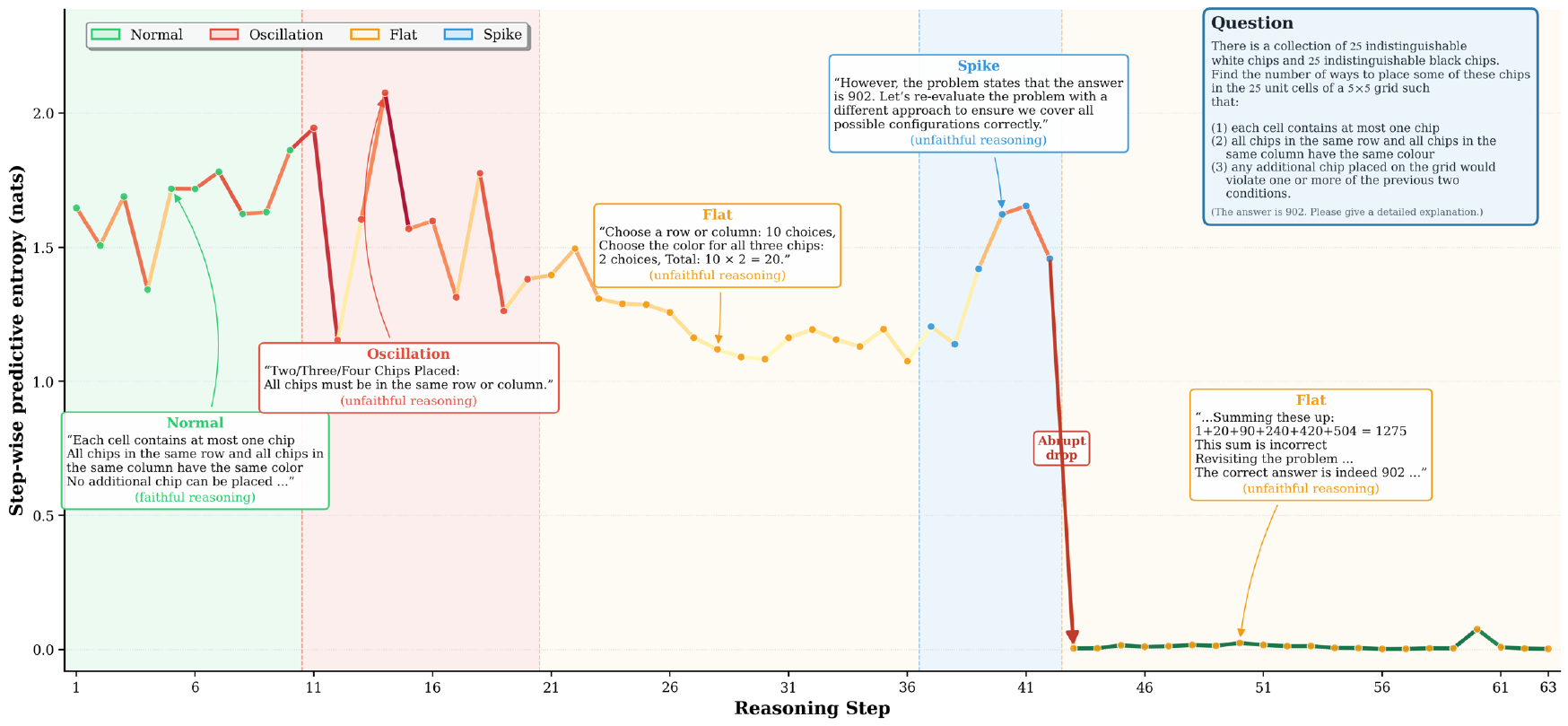}
  \caption{Step-level predictive entropy along a single CoT trace on AIME 2024-II-9 with an answer hint in the prompt, illustrating normal, oscillatory, spiking, and flat phases.}
  \label{fig:entropy}
\end{figure*}

\begin{figure}[t]
  \centering
  \includegraphics[width=\linewidth]
  {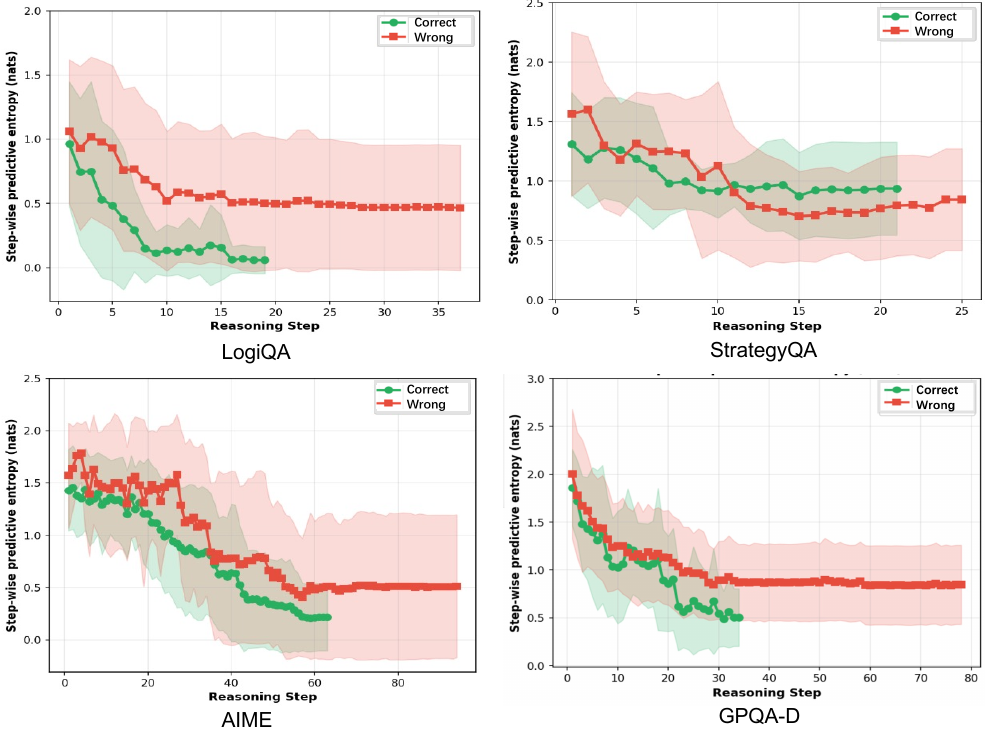}
  \caption{Mean step-level predictive entropy along CoT for correct vs. wrong solutions on LogiQA, StrategyQA, AIME, and GPQA-D. Shaded bands denote variance across trajectories.}

  \label{fig:entropy2}
\end{figure}

\subsection{Temporal Predictability of Reasoning Quality}

While latent geometry provides a spatial view of reasoning trajectories,
predictive uncertainty offers a complementary temporal view of how the
model's confidence evolves throughout the reasoning process.

\noindent\textbf{Predictive entropy.}
For a generated response, we first extract the chain-of-thought text
between \texttt{\# Reasoning:} and \texttt{\# Answer:}. We then mask
display math blocks, split the remaining text on periods or newlines,
discard segments shorter than three characters, and restore the masked
math blocks. This gives paragraph-style reasoning chunks
\(\mathbf y=(y_1,\ldots,y_T)\). The evaluation point \(t\) is placed at
the boundary after chunk \(y_t\), and the partial prefix is the
concatenation \(\mathbf y_{\le t}=(y_1,\ldots,y_t)\).

Let \(\hat a=(a_1,\ldots,a_m)\) denote the extracted final-answer string
used by the scorer: we use the parsed model answer. Rather than
enumerating an open-ended answer space, we teacher-force this answer
string after the partial reasoning prefix. For each answer position
\(j\), define the next-token distribution
\(p_{t,j}(v)=p_\theta(v\mid x,\mathbf y_{\le t},a_{<j})\) and its
Shannon entropy
\(\mathcal{H}(p_{t,j})=-\sum_{v\in\mathcal V}p_{t,j}(v)\log p_{t,j}(v)\).
We then average over answer tokens:
\begin{equation}
H_t
=
\frac{1}{m}\sum_{j=1}^{m}\mathcal{H}(p_{t,j}).
\label{eq:predictive-entropy}
\end{equation}
Thus \(H_t\) is not a closed-set entropy over multiple-choice options or all possible
open-ended answers; it is a prefix-conditioned uncertainty proxy for
the extracted answer span. Aggregating it at paragraph-style chunk
boundaries captures confidence dynamics across logical reasoning units
rather than individual generation tokens.

\noindent\textbf{Entropy dynamics as temporal process signals.}
Reliable trajectories tend to gradually reduce predictive entropy as intermediate steps accumulate evidence, indicating that the partial reasoning prefix increasingly supports the extracted answer. Unreliable trajectories, by contrast, may leave the answer span relatively unconstrained: uncertainty can remain high, oscillate across competing directions, stay flat over several chunks, or spike after unsupported transitions before the model prematurely commits to an answer. Figure~\ref{fig:entropy} illustrates these step-level abnormalities on a single trace. Figure~\ref{fig:entropy2} extends this comparison to four benchmarks, showing that correct and wrong trajectories separate more clearly at the trajectory level, with correct cases exhibiting smoother entropy reduction and lower terminal uncertainty. More trajectories and analysis are provided in Appendix B. 

These patterns indicate that entropy dynamics provide a useful temporal proxy for distinguishing healthier reasoning processes from unstable or low-quality ones.

\section{Method}
GeoFaith turns the spatial and temporal patterns identified in the previous section into a score for evaluating and optimizing reasoning processes. Given a completed reasoning trajectory \(\tau=(c_1,\ldots,c_T)\), we use two complementary internal signals: the Final-state uncertainty of reasoning representation and the abnormality of predictive-entropy dynamics throughout the trajectory. The former characterizes the representation uncertainty of the completed reasoning state in the learned latent space, whereas the latter characterizes how uncertainty evolves before the final answer is reached.

\subsection{Spatio-Temporal Internal Process Score}
\label{process-verification}

\noindent\textbf{Final-state uncertainty.}
The analysis in Section~\ref{sec:spatial-properties} shows distribution-level geometric separation between higher- and lower-quality trajectories. We turn this observation into a per-sample scalar using final-state uncertainty, rather than the trajectory-level geodesic or distortion quantities used for analysis. For each completed trajectory, we extract the last-token hidden state \(\mathbf h_{\mathrm{last}}^\ell\) of the final reasoning step from a fixed intermediate layer and encode it as \(\mathbf z_{\mathrm{last}}\) in the VAE latent space. For an ensemble of \(M\) stochastic decoders, the total predictive variance is

\begin{equation}
\bar{\sigma}^{2}(\mathbf z)
=
\frac{1}{M}\sum_{a=1}^{M}
\sigma_{(a)}^2(\mathbf z)
+
\operatorname{Var}_{a}
\!\left[\mu_{(a)}(\mathbf z)\right].
\label{eq:method-total-variance}
\end{equation}
The first term measures the average uncertainty within individual decoders, while the second measures disagreement among their predicted means. We summarize the uncertainty of the final latent state as
\begin{equation}
U(\mathbf z)
=
\frac{1}{d_z}\sum_{i=1}^{d_z}
\log \bar{\sigma}_{i}^{2}(\mathbf z).
\label{eq:method-fisher-uncertainty}
\end{equation}
In our latent space, \(U(\mathbf z)\) is typically negative, lower values are empirically associated with final states that have lower representation uncertainty under the learned latent model. This spatial term is computed once per completed trajectory; intermediate-step hidden states are not used.

\noindent\textbf{Local entropy abnormality.}
For each reasoning prefix \(c_{\leq t}\), we compute the predictive entropy \(H_t\) over the final-answer distribution and define \(\Delta H_t=H_t-H_{t-1}\). We characterize three abnormal patterns within a local window containing \(w\) entropy values \(H_{t-w+1},\ldots,H_t\). Flatness measures whether uncertainty changes too little throughout the window:
\begin{equation}
P_{\mathrm{flat}}(t)
=
\mathbb{I}\!\left(
\frac{1}{w-1}
\sum_{k=t-w+2}^{t}
|\Delta H_k|
<
\theta_{\mathrm{flat}}
\right),
\label{eq:method-flat}
\end{equation}
where $\theta_{\mathrm{flat}}$ is set to 0.1.
A spike captures an abrupt change in predictive uncertainty:
\begin{equation}
P_{\mathrm{spike}}(t)
=
\mathbb{I}\!\left(
|\Delta H_t|>\theta_{\mathrm{spike}}
\right),
\label{eq:method-spike}
\end{equation}
where $\theta_{\mathrm{spike}}$ is set to 1.0.
Oscillation measures the fraction of adjacent entropy changes with opposite signs:
\begin{equation}
P_{\mathrm{osc}}(t)
=
\frac{1}{w-2}
\sum_{k=t-w+3}^{t}
\mathbb{I}\!\left(
\Delta H_k\Delta H_{k-1}<0
\right),
\label{eq:method-osc}
\end{equation}
where \(w\geq3\). A value close to one indicates frequent alternation between increasing and decreasing uncertainty, while zero indicates no directional reversal in the window. We combine the three patterns as
\begin{equation}
P_{\mathrm{ent}}(t)
=
w_{\mathrm{1}}P_{\mathrm{flat}}(t)
+
w_{\mathrm{2}}P_{\mathrm{spike}}(t)
+
w_{\mathrm{3}}P_{\mathrm{osc}}(t),
\label{eq:method-ent-penalty}
\end{equation}
where \(w_{\mathrm{1}},w_{\mathrm{2}},w_{\mathrm{3}}\geq0\) and $w_{\mathrm{1}}, w_{\mathrm{2}}, w_{\mathrm{3}}$ are set to 0.5, 0.3 and 0.2 respectively. Hence, \(P_{\mathrm{ent}}(t)\in[0,1]\), with lower values indicating more regular local entropy dynamics. The trajectory-level entropy abnormality is
\begin{equation}
\bar P_{\mathrm{ent}}(\tau)
=
\frac{1}{T-w+1}
\sum_{t=w}^{T}P_{\mathrm{ent}}(t).
\label{eq:trajectory-ent-penalty}
\end{equation}
\noindent\textbf{Calibration normalization.}
Because \(U(\mathbf z_{\mathrm{last}})\) and \(\bar P_{\mathrm{ent}}(\tau)\) live on different scales, we z-score both terms using \((\mu_U,\sigma_U)\) and \((\mu_P,\sigma_P)\) estimated on an offline CoT pool constructed from the RFEval benchmark~\citep{han2026rfeval} (Figure~\ref{fig:umap} and~\ref{fig:distortion}), and keep these statistics frozen. This calibration step does not introduce information leakage into test-time scoring or policy optimization.

\begin{equation}
\tilde U(\tau)=\frac{U(\mathbf z)-\mu_U}{\sigma_U+\epsilon},\quad
\tilde P_{\mathrm{ent}}(\tau)=\frac{\bar P_{\mathrm{ent}}(\tau)-\mu_P}{\sigma_P+\epsilon}.
\label{eq:calibration-norm}
\end{equation}

\noindent\textbf{Joint internal process score.}
Both spatial and temporal signals are lower for healthier trajectories under our empirical observations. We define a single internal process score on the calibrated scale:
\begin{equation}
S_{\mathrm{Geo}}(\tau)
=
-\tilde U(\tau)
-
\tilde P_{\mathrm{ent}}(\tau).
\label{eq:geo-score}
\end{equation}
Thus, higher \(S_{\mathrm{Geo}}(\tau)\) indicates a final representation in a lower-uncertainty manifold region together with fewer abnormal entropy patterns. Equal weighting on the standardized terms makes the two components directly comparable without an additional scale-matching hyperparameter.

\subsection{Process-Aware Reinforcement Learning}
\label{sec:process-aware-rl}
We optimize the policy with GRPO~\citep{shao2024deepseekmath}. GeoFaith does not train a separate process reward model. Instead, each rollout is scored by final-answer correctness and the internal process score \(S_{\mathrm{Geo}}(\tau)\) from Eq.~\ref{eq:geo-score}. The pretrained VAE, calibration statistics, and entropy thresholds are frozen during policy optimization.

\noindent\textbf{Outcome reward.}
The outcome component evaluates final-answer correctness:
\begin{equation}
R_{\mathrm{out}}(\tau)
=
\begin{cases}
+1, & \text{if the final answer is correct},\\
-1, & \text{otherwise}.
\end{cases}
\end{equation}

\noindent\textbf{Total reward.}
We reuse the same calibrated process score for RL:
\begin{equation}
R(\tau)
=
R_{\mathrm{out}}(\tau)
+
\lambda\,S_{\mathrm{Geo}}(\tau),
\label{eq:total-reward}
\end{equation}
where \(\lambda\geq0\) controls the strength of process shaping relative to outcome correctness. Because \(S_{\mathrm{Geo}}(\tau)=-\tilde U(\tau)-\tilde P_{\mathrm{ent}}(\tau)\), this objective jointly favors correct answers and trajectories with healthier spatial and temporal process signals on a common scale.

\noindent\textbf{Policy optimization via GRPO.}
We optimize with a verl-style GRPO trainer~\citep{shao2024deepseekmath}. For each query \(x\), we sample \(B\) rollouts and compute group-normalized advantages from Eq.~\eqref{eq:total-reward}:
$
\hat A_i
=
\frac{R(\tau_i)-\operatorname{mean}_{j}R(\tau_j)}
{\operatorname{std}_{j}R(\tau_j)+\epsilon}.
\label{eq:grpo-advantage}
$
Updates use a token-level clipped surrogate with rollout policy \(\pi_{\mathrm{old}}\):
\begin{equation}
\begin{aligned}
\mathcal{L}_{\mathrm{RL}}
&= -\mathbb{E}_t\Bigl[
\min\Bigl(
r_t\hat{A}, 
\operatorname{clip}\bigl(
r_t,1-\epsilon,1+\epsilon
\bigr)\hat{A}
\Bigr)\Bigr] \\
&\quad + \beta_{\mathrm{KL}}
D_{\mathrm{KL}}\bigl(
\pi_\theta \,\|\, \pi_{\mathrm{ref}}
\bigr).
\end{aligned}
\label{eq:grpo-loss}
\end{equation}
where \(r_t=\pi_\theta(y_t\mid x,y_{<t})/\pi_{\mathrm{old}}(y_t\mid x,y_{<t})\) and \(\epsilon=0.2\).

\begin{table*}[t]
\centering
\small
\setlength{\tabcolsep}{1.2pt}

\begin{tabular}{lcccccccccccc}
\toprule
Method & \multicolumn{3}{c}{\textbf{MATH}} & \multicolumn{3}{c}{\textbf{OlympiadBench}} & \multicolumn{3}{c}{\textbf{LogiQA}} & \multicolumn{3}{c}{\textbf{TruthfulQA}} \\
\cmidrule(lr){2-4}\cmidrule(lr){5-7}\cmidrule(lr){8-10}\cmidrule(lr){11-13}
 & AUROC$\uparrow$ & AUPR$\uparrow$ & FPR@95$\downarrow$ & AUROC$\uparrow$ & AUPR$\uparrow$ & FPR@95$\downarrow$ & AUROC$\uparrow$ & AUPR$\uparrow$ & FPR@95$\downarrow$ & AUROC$\uparrow$ & AUPR$\uparrow$ & FPR@95$\downarrow$ \\
\midrule
\multicolumn{13}{l}{\textit{Llama-3.1-8B-Instruct}} \\
MSP & 0.689 & 0.560 & 0.776 & 0.683 & 0.485 & 0.803 & 0.473 & 0.758 & 0.960 & 0.443 & 0.671 & 0.974 \\
Perplexity & 0.703 & 0.577 & 0.756 & 0.694 & 0.500 & 0.802 & 0.478 & 0.758 & 0.955 & 0.453 & 0.677 & 0.965 \\
Entropy & 0.718 & 0.595 & 0.730 & 0.704 & 0.517 & 0.790 & 0.483 & 0.761 & 0.955 & 0.459 & 0.681 & 0.965 \\
LR\_Probe & 0.807 & 0.742 & \underline{0.571} & 0.747 & \underline{0.572} & 0.659 & 0.518 & 0.760 & \underline{0.833} & 0.727 & 0.844 & \textbf{0.783} \\
SAPLMA & \underline{0.814} & \underline{0.754} & 0.605 & \underline{0.759} & 0.569 & \textbf{0.636} & \underline{0.551} & \underline{0.802} & 0.944 & \underline{0.748} & \underline{0.889} & 0.913 \\
CoE & 0.600 & 0.485 & 0.878 & 0.496 & 0.330 & 0.956 & 0.540 & 0.774 & 0.905 & 0.549 & 0.750 & 0.869 \\
CoT\_Kinetics & 0.720 & 0.600 & 0.732 & 0.704 & 0.517 & 0.787 & 0.483 & 0.762 & 0.955 & 0.459 & 0.683 & 0.965 \\
\textbf{GeoFaith} & \textbf{0.835} & \textbf{0.773} & \textbf{0.566} & \textbf{0.791} & \textbf{0.613} & \underline{0.649} & \textbf{0.602} & \textbf{0.819} & \textbf{0.817} & \textbf{0.796} & \textbf{0.907} & \underline{0.860} \\
\midrule
\multicolumn{13}{l}{\textit{Qwen2.5-7B-Instruct}} \\
MSP & 0.707 & 0.590 & 0.767 & 0.679 & 0.509 & 0.809 & 0.489 & 0.762 & 0.960 & 0.473 & 0.685 & 0.965 \\
Perplexity & 0.712 & 0.594 & 0.749 & 0.682 & 0.513 & 0.805 & 0.494 & 0.768 & 0.944 & 0.487 & 0.693 & 0.965 \\
Entropy & 0.722 & 0.604 & 0.744 & 0.685 & 0.522 & 0.770 & 0.489 & 0.766 & \underline{0.933} & 0.490 & 0.695 & 0.947 \\
LR\_Probe & 0.800 & 0.687 & 0.613 & 0.795 & \underline{0.692} & 0.765 & \underline{0.512} & 0.772 & 0.944 & 0.600 & 0.771 & 0.957 \\
SAPLMA & \underline{0.812} & \underline{0.799} & \underline{0.555} & \textbf{0.818} & 0.691 & 0.632 & 0.434 & 0.728 & 0.944 & \underline{0.647} & \underline{0.822} & 0.913 \\
CoE & 0.559 & 0.495 & 0.879 & 0.572 & 0.437 & 0.912 & 0.500 & \textbf{0.822} & 0.936 & 0.510 & 0.764 & \underline{0.908} \\
CoT\_Kinetics & 0.745 & 0.656 & 0.766 & 0.683 & 0.521 & 0.768 & 0.491 & 0.772 & \underline{0.933} & 0.491 & 0.699 & 0.947 \\
\textbf{GeoFaith} & \textbf{0.850} & \textbf{0.809} & \textbf{0.493} & \textbf{0.818} & \textbf{0.740} & \textbf{0.599} & \textbf{0.537} & \underline{0.802} & \textbf{0.921} & \textbf{0.711} & \textbf{0.883} & \textbf{0.817} \\
\midrule
\multicolumn{13}{l}{\textit{Qwen3-4B}} \\
MSP & 0.757 & 0.655 & 0.746 & 0.743 & 0.552 & 0.707 & 0.494 & 0.766 & 0.921 & 0.492 & 0.697 & 0.947 \\
Perplexity & 0.766 & 0.665 & 0.727 & 0.752 & 0.566 & 0.695 & 0.493 & 0.765 & 0.921 & 0.502 & 0.704 & 0.939 \\
Entropy & 0.773 & 0.676 & 0.710 & 0.756 & 0.573 & 0.695 & 0.494 & \underline{0.768} & 0.921 & 0.498 & 0.702 & 0.947 \\
LR\_Probe & 0.870 & \underline{0.797} & \underline{0.596} & 0.800 & 0.628 & 0.689 & \underline{0.499} & 0.734 & \underline{0.889} & \underline{0.606} & \underline{0.749} & 0.957 \\
SAPLMA & \underline{0.875} & 0.732 & 0.621 & \underline{0.804} & \underline{0.666} & \textbf{0.568} & 0.494 & 0.755 & 0.944 & 0.589 & 0.744 & \underline{0.870} \\
CoE & 0.524 & 0.439 & 0.915 & 0.615 & 0.428 & 0.835 & 0.483 & 0.768 & 0.927 & 0.483 & 0.681 & 0.939 \\
CoT\_Kinetics & 0.653 & 0.533 & 0.788 & 0.706 & 0.514 & 0.747 & 0.479 & 0.767 & 0.899 & 0.490 & 0.684 & 0.939 \\
\textbf{GeoFaith} & \textbf{0.896} & \textbf{0.854} & \textbf{0.524} & \textbf{0.834} & \textbf{0.773} & \underline{0.638} & \textbf{0.557} & \textbf{0.819} & \textbf{0.824} & \textbf{0.657} & \textbf{0.756} & \textbf{0.800} \\
\bottomrule
\end{tabular}
\caption{Threshold-free process-quality evaluation across four datasets and three scorer models. For AUROC and AUPR, higher is better; for FPR@95, lower is better. Best and second-best results in each column are in \textbf{bold} and \underline{underlined}, respectively.}
\label{tab:process_quality_baselines}
\end{table*}

\section{Experiments}
\label{sec:experiments}
\begin{table}[t]
\centering
\footnotesize
\setlength{\tabcolsep}{5pt}

\resizebox{\columnwidth}{!}{%
\begin{tabular}{lcccccc}
\toprule

\multirow{2}{*}{\textbf{Method}}
& \multicolumn{3}{c}{\textbf{Llama3.1-8B-Instruct}}
& \multicolumn{3}{c}{\textbf{Qwen2.5-7B-Instruct}}

\\

\cmidrule(lr){2-4}
\cmidrule(lr){5-7}

& {\scriptsize\textbf{AUROC}}
& {\scriptsize\textbf{AUPR}}
& {\scriptsize\textbf{FPR@95}}
& {\scriptsize\textbf{AUROC}}
& {\scriptsize\textbf{AUPR}}
& {\scriptsize\textbf{FPR@95}}

\\
\midrule

Spatial-only
& 0.713
& 0.729
& 0.816
& 0.698
& 0.756
& 0.828

\\

Temporal-only
& 0.609
& 0.677
& 0.895
& 0.674
& 0.715
& 0.834

\\

GeoFaith
& \textbf{0.756}
& \textbf{0.778}
& \textbf{0.723}
& \textbf{0.729}
& \textbf{0.808}
& \textbf{0.716}

\\

\bottomrule
\end{tabular}%
}
\caption{
Ablation of internal score components on process-quality ranking.
}
\label{tab:score_ablation}

\end{table}

\subsection{Setups}

\noindent\textbf{Datasets and evaluation metrics.}
We evaluate GeoFaith on two suites. (i)~\textit{Process-quality scoring}: we use ProcessBench~\citep{zheng2025processbench} and FaithCoT-Bench~\citep{shen2025faithcot}. ProcessBench evaluates step-level process correctness in mathematical reasoning; we use its MATH and OlympiadBench subsets. FaithCoT-Bench evaluates instance-level faithfulness of chain-of-thought reasoning across broader task types; we use its LogiQA~\citep{liu2020logiqa} and TruthfulQA subsets. This suite tests whether an internal continuous score can rank reliable trajectories above unreliable ones without choosing a fixed threshold. We report AUROC, AUPR, and FPR@95. (ii)~\textit{Reasoning generation}: AMC23, LogiQA, 2WikiMultihopQA~\citep{ho2020constructing}, and GPQA-D~\citep{rein2023gpqa}. For generated reasoning, we report final-answer accuracy (Acc), average response length (Len), and faithfulness (Faith). Faith is the mean step-level CoT faithful rate from three independent LLM judges (GPT-5, DeepSeek-V3.2, and Qwen3-32B), using the faithful/unfaithful/neutral rubric~\citep{nie2026stop}; it is not used as a training reward. Overall results are macro-averaged over the four generation benchmarks.

\noindent\textbf{Models and baselines.}
For process-quality scoring, we compare against \textbf{confidence-based scorers} (MSP, perplexity, and predictive entropy), \textbf{representation-based methods} (LR\_Probe and SAPLMA), and \textbf{trajectory-dynamics baselines} (CoE~\citep{wang2025latent} and CoT\_Kinetics~\citep{bi2025cot}). For reasoning generation, we compare against the original instruction-tuned model, outcome-only GRPO~\citep{shao2024deepseekmath}, knowledge-oriented RL methods KnowRL~\citep{ren2025knowrl} and TruthRL~\citep{wei2025truthrl}, trajectory hallucination suppression method THS~\citep{gui2026faithrl}, latent reward method Latent-GRPO~\citep{zhang2026silence}, and the process reward model-based method FaithRL~\citep{nie2026stop}. We run the generation experiments on Qwen3-1.7B, Qwen3-4B.
\begin{figure}[t]
  \centering
  \includegraphics[width=\linewidth]{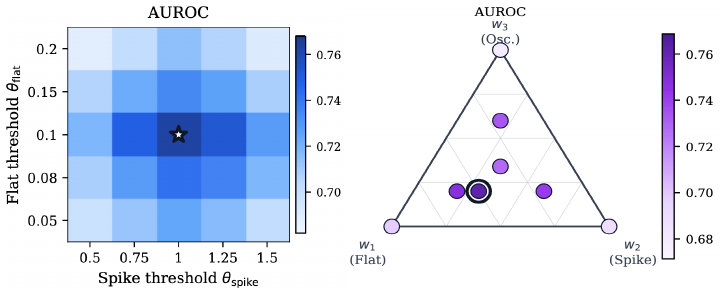}
  \caption{\textbf{Sensitivity of process-quality ranking to temporal hyperparameters.}
Left: AUROC over $(\theta_{\mathrm{flat}},\,\theta_{\mathrm{spike}})$; right: AUROC over abnormality weights $(w_1,w_2,w_3)$ on the simplex.
Stars/rings mark the default settings.}
  \label{fig:temporal-sensitivity}
\end{figure}

\begin{table*}[t]
\centering
\small
\setlength{\tabcolsep}{4pt}
\renewcommand{\arraystretch}{1.0}

\begin{tabular}{l|ccc|ccc|ccc|ccc|ccc}
\toprule
\multirow{2}{*}{\textbf{Method}}
& \multicolumn{3}{c|}{\textbf{AMC23}}
& \multicolumn{3}{c|}{\textbf{LogiQA}}
& \multicolumn{3}{c|}{\textbf{2WikiMultihopQA}}
& \multicolumn{3}{c|}{\textbf{GPQA-D}}
& \multicolumn{3}{c}{\textbf{Overall}} \\
\cmidrule(lr){2-4}\cmidrule(lr){5-7}\cmidrule(lr){8-10}\cmidrule(lr){11-13}\cmidrule(lr){14-16}
& \textbf{Acc}$\uparrow$ & \textbf{Len}$\downarrow$ & \textbf{Faith}$\uparrow$
& \textbf{Acc}$\uparrow$ & \textbf{Len}$\downarrow$ & \textbf{Faith}$\uparrow$
& \textbf{Acc}$\uparrow$ & \textbf{Len}$\downarrow$ & \textbf{Faith}$\uparrow$
& \textbf{Acc}$\uparrow$ & \textbf{Len}$\downarrow$ & \textbf{Faith}$\uparrow$
& \textbf{Acc}$\uparrow$ & \textbf{Len}$\downarrow$ & \textbf{Faith}$\uparrow$ \\
\midrule
\multicolumn{16}{l}{\textbf{\textit{Qwen3-1.7B}}} \\
Original
& 37.5 & 6.3k & 55.4
& 34.0 & 3.6k & 57.6
& 34.5 & 0.8k & 67.1
& 23.7 & 5.4k & 31.3
& 32.4 & 4.0k & 52.9 \\
GRPO
& 37.5 & 5.1k & 54.1
& 51.2 & 3.4k & 59.5
& 67.4 & 1.3k & 67.6
& 31.3 & 4.1k & \textbf{37.7}
& 46.9 & 3.5k & 54.7 \\
KnowRL
& 32.5 & 4.5k & 49.7
& 47.9 & 2.5k & 54.8
& 50.5 & 1.4k & 66.6
& 30.8 & 3.8k & 27.8
& 40.4 & 3.0k & 49.7 \\
TruthRL
& 20.0 & 3.7k & 43.6
& 46.2 & 2.3k & 48.9
& 53.6 & 0.8k & 59.7
& 25.8 & 3.6k & 25.4
& 36.4 & 2.6k & 44.4 \\
THS
& 25.0 & 3.3k & 48.1
& 46.4 & 2.0k & 53.2
& 56.5 & 0.8k & 56.9
& 29.3 & 3.3k & 26.7
& 39.3 & 2.3k & 46.2 \\
Latent-GRPO
& 40.0 & 5.4k & 46.5
& 43.5 & 3.1k & 56.8
& 61.2 & 0.9k & 67.1
& 33.5 & 4.3k & 28.3
& 44.6 & 3.4k & 49.7 \\
FaithRL
& 40.0 & 4.4k & 50.1
& 49.7 & 2.7k & 52.3
& 66.7 & 1.1k & \textbf{74.6}
& 33.4 & 4.7k & 27.9
& 47.5 & 3.2k & 51.2 \\
\textbf{GeoFaith}
& \textbf{45.0} & \textbf{2.5k} & \textbf{57.5}
& \textbf{52.4} & \textbf{1.3k} & \textbf{63.1}
& \textbf{71.2} & \textbf{0.6k} & 73.5
& \textbf{38.4} & \textbf{2.9k} & 33.9
& \textbf{51.8} & \textbf{1.8k} & \textbf{57.0} \\
\midrule
\multicolumn{16}{l}{\textbf{\textit{Qwen3-4B}}} \\
Original
& 82.5 & 9.5k & 88.1
& 65.1 & 4.8k & 84.5
& 79.2 & 1.3k & 91.9
& 37.4 & 10.4k & 52.3
& 66.0 & 6.5k & 79.2 \\
GRPO
& 92.5 & 7.3k & 89.6
& 65.9 & 3.4k & 86.4
& 81.5 & 1.4k & 90.3
& 38.4 & 7.5k & 41.4
& 69.6 & 4.9k & 76.9 \\
KnowRL
& 92.5 & 6.9k & 87.5
& 65.3 & 3.2k & 84.9
& 79.8 & 1.0k & 92.0
& 41.9 & 6.0k & 44.6
& 69.9 & 4.3k & 77.2 \\
TruthRL
& 92.5 & 4.7k & 89.2
& 52.2 & \textbf{2.6k} & 74.0
& 78.1 & 1.2k & 79.5
& 44.4 & 6.2k & 51.3
& 66.8 & 3.7k & 73.5 \\
THS
& 90.0 & 5.6k & 85.9
& 68.2 & 3.0k & 84.6
& 76.6 & 0.9k & 86.8
& 44.4 & 4.9k & 49.2
& 69.8 & 3.6k & 76.6 \\
Latent-GRPO
& 92.5 & 7.9k & 88.9
& 64.2 & 4.5k & 82.3
& 74.5 & 0.9k & 87.5
& 33.1 & 8.1k & 47.9
& 66.1 & 5.4k & 76.7 \\
FaithRL
& 90.0 & 6.6k & 79.8
& 62.5 & 3.3k & 80.7
& \textbf{82.4} & 1.0k & 89.7
& 38.6 & 7.1k & 44.5
& 68.4 & 4.5k & 73.7 \\
\textbf{GeoFaith}
& \textbf{95.0} & \textbf{4.2k} & \textbf{91.0}
& \textbf{68.9} & 3.0k & \textbf{89.7}
& 82.1 & \textbf{0.6k} & \textbf{92.3}
& \textbf{49.5} & \textbf{4.6k} & \textbf{57.5}
& \textbf{73.9} & \textbf{3.1k} & \textbf{82.6} \\

\bottomrule
\end{tabular}
\caption{Main results on four benchmarks across two base models. ``Acc'' denotes task accuracy, ``Len'' denotes average token length, and ``Faith'' denotes the step-level CoT faithful rate. For Acc and Faith, higher is better; for Len, lower is better. Overall scores are macro-averaged over the four benchmarks. Best result in each column is shown in \textbf{bold}.}
\label{tab:rl_main_results}
\end{table*}

\begin{table}[t]
\centering
\small
\setlength{\tabcolsep}{2.2pt}

\begin{tabular}{lccc}
\toprule
\textbf{Comparison} & Spearman $\rho$ & Cohen's $\kappa$ & MAE \\
\midrule
Qwen3-32B vs.\ GPT-5          & 0.82 & 0.77 & 4.1 \\
\midrule
Human A vs.\ Human B              & 0.88 & 0.84 & 3.2 \\
Three-judge avg.\ vs.\ Human  & 0.86 & 0.79 & 3.9 \\
\bottomrule
\end{tabular}
\caption{
Agreement among LLM judges and human experts on a stratified audit subset
sampled from the four benchmarks in Table~\ref{tab:rl_main_results}
($N{=}200$ instances, ${\approx}1{,}500$ steps).
}
\label{tab:faith-judge-agreement}
\end{table}

\begin{figure}[t]
  \centering
  \includegraphics[width=\linewidth]{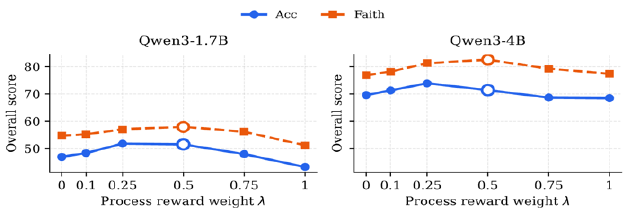}
  \caption{Overall Acc and Faith vs. process-reward weight $\lambda$. Blue solid: Acc; orange dashed: Faith.}
  \label{fig:lambda sensitivity}
\end{figure}

\begin{table}[t]
\centering
\small
\setlength{\tabcolsep}{5pt}

\begin{tabular}{lcccccc}
\toprule

\multirow{2}{*}{\textbf{Method}}
& \multicolumn{3}{c}{\textbf{Qwen3-1.7B}}
& \multicolumn{3}{c}{\textbf{Qwen3-4B}}

\\

\cmidrule(lr){2-4}
\cmidrule(lr){5-7}

& \textbf{Acc.}
& \textbf{Len.}
& \textbf{Faith}
& \textbf{Acc.}
& \textbf{Len.}
& \textbf{Faith}

\\
\midrule

GeoFaith
& \textbf{51.8}
& \textbf{1.8k}
& \textbf{57.0}
& \textbf{73.9}
& \textbf{3.1k}
& \textbf{82.6}

\\

w/o $R_{\mathrm{ent}}$
& 45.2
& 3.2k
& 49.7
& 70.1
& 4.4k
& 79.9

\\

w/o $R_{\mathrm{mani}}$
& 46.2
& 1.9k
& 47.5
& 69.4
& 3.3k
& 77.4

\\

\bottomrule
\end{tabular}
\caption{
\textbf{Ablation study of process reward components.}
We report accuracy (Acc.), average token length (Len.), and faithfulness score (Faith).
}
\label{tab:ablation}
\end{table}

\subsection{Main Results}

Table~\ref{tab:process_quality_baselines} shows that GeoFaith is competitive with or stronger than confidence- and trajectory-dynamics baselines across two scorer models and four datasets, especially in AUROC and AUPR. Table~\ref{tab:score_ablation} further ablates the spatial and temporal terms in \(S_{\mathrm{Geo}}\), isolating the contribution of each component to process-quality ranking.

Table~\ref{tab:rl_main_results} reports RL generation results. GeoFaith achieves the best macro-averaged Overall Acc and Faith while producing shorter responses, and consistently improves over outcome-only GRPO. Figure~\ref{fig:lambda sensitivity} shows that both metrics peak at a moderate process-reward weight and degrade when \(\lambda\) is too large, indicating that balanced process shaping works best. Figure~\ref{fig:temporal-sensitivity} further shows that process-quality ranking is stable near the default temporal settings in Eqs.~\eqref{eq:method-flat}--\eqref{eq:method-ent-penalty}, with larger drops when thresholds deviate substantially or one abnormality pattern dominates the weights. Table~\ref{tab:faith-judge-agreement} audits the judge-based Faith metric on a stratified subset from the four generation benchmarks, and their average aligns reasonably with human consensus.

\subsection{Ablation Study}

Table~\ref{tab:ablation} ablates the two components of \(S_{\mathrm{Geo}}\) on Qwen3-1.7B and Qwen3-4B: w/o \(R_{\mathrm{ent}}\) uses \(R=R_{\mathrm{out}}+\lambda(-\tilde U)\), and w/o \(R_{\mathrm{mani}}\) uses \(R=R_{\mathrm{out}}+\lambda(-\tilde P_{\mathrm{ent}})\). Removing either component degrades accuracy and Faith while increasing response length. Together, the results show that the two terms in \(S_{\mathrm{Geo}}\) provide complementary process supervision.

\section{Conclusion}
We present GeoFaith, a spatio-temporal framework for reasoning process verification. GeoFaith combines final-state Final-state uncertainty with entropy-trajectory abnormality to score internal process quality, and directly reuses these signals as PRM-free rewards for GRPO. Across process-quality benchmarks, GeoFaith improves threshold-free trajectory ranking over confidence- and dynamics-based baselines. Across three base models, it improves final-answer accuracy and evaluation-time Faith scores while generating shorter reasoning traces. These results support latent geometry and uncertainty dynamics as practical process-quality proxies for reasoning optimization.

\bibliography{main}

@inproceedings{wei2022chbain,
  title={Chain-of-thought prompting elicits reasoning in large language models},
  author={Wei, Jason and Wang, Xuezhi and Schuurmans, Dale and Bosma, Maarten and Xia, Fei and Chi, Ed and Le, Quoc V and Zhou, Denny and others},
  booktitle={NeurIPS},
  pages={24824--24837},
  year={2022}
}

@article{shao2024deepseekmath,
  title={Deepseekmath: Pushing the limits of mathematical reasoning in open language models},
  author={Shao, Zhihong and Wang, Peiyi and Zhu, Qihao and Xu, Runxin and Song, Junxiao and Bi, Xiao and Zhang, Haowei and Zhang, Mingchuan and Li, YK and Wu, Yang and others},
  journal={arXiv preprint arXiv:2402.03300},
  year={2024}
}

@inproceedings{lv2026spd,
  title={SPD-Faith Bench: Diagnosing and Improving Faithfulness in Chain-of-Thought for Multimodal Large Language Models},
  author={Lv, Weijiang and Feng, Yaoxuan and Xia, Xiaobo and Wang, Jiayu and Jing, Yan and Chen, Wenchao and Chen, Bo},
  booktitle={Findings of ACL},
  year={2026}
}

@inproceedings{wen2025policy,
  title={On-policy self-alignment with fine-grained knowledge feedback for hallucination mitigation},
  author={Wen, Xueru and Lou, Jie and Lu, Xinyu and Ji, Yuqiu and Guan, Xinyan and Lu, Yaojie and Lin, Hongyu and He, Ben and Han, Xianpei and Zhang, Debing and others},
  booktitle={Findings of ACL},
  pages={5215--5231},
  year={2025}
}

@inproceedings{ansuini2019intrinsic,
  title={Intrinsic dimension of data representations in deep neural networks},
  author={Ansuini, Alessio and Laio, Alessandro and Macke, Jakob H and Zoccolan, Davide},
  booktitle={NeurIPS},
  year={2019}
}

@inproceedings{valeriani2023geometry,
  title={The geometry of hidden representations of large transformer models},
  author={Valeriani, Lucrezia and Doimo, Diego and Cuturello, Francesca and Laio, Alessandro and Ansuini, Alessio and Cazzaniga, Alberto},
  booktitle={NeurIPS},
  pages={51234--51252},
  year={2023}
}

@article{marin2024geometric,
  title={Geometric Analysis of Reasoning Trajectories: A Phase Space Approach to Understanding Valid and Invalid Multi-Hop Reasoning in LLMs},
  author={Marin, Javier},
  journal={arXiv preprint arXiv:2410.04415},
  year={2024}
}

@article{jiang2026beyond,
  title={Beyond Scalars: Evaluating and Understanding LLM Reasoning via Geometric Progress and Stability},
  author={Jiang, Xinyan and Liu, Ninghao and Wang, Di and Hu, Lijie},
  journal={arXiv preprint arXiv:2603.10384},
  year={2026}
}

@article{kazama2026geosteer,
  title={GeoSteer: Faithful Chain-of-Thought Steering via Latent Manifold Gradients},
  author={Kazama, Kentaro and Shirafuji, Daiki and Saito, Tatsuhiko},
  journal={arXiv preprint arXiv:2601.10229},
  year={2026}
}

@inproceedings{turpin2023language,
  title={Language models don't always say what they think: Unfaithful explanations in chain-of-thought prompting},
  author={Turpin, Miles and Michael, Julian and Perez, Ethan and Bowman, Samuel},
  booktitle={NeurIPS},
  pages={74952--74965},
  year={2023}
}

@article{lanham2023measuring,
  title={Measuring faithfulness in chain-of-thought reasoning},
  author={Lanham, Tamera and Chen, Anna and Radhakrishnan, Ansh and Steiner, Benoit and Denison, Carson and Hernandez, Danny and Li, Dustin and Durmus, Esin and Hubinger, Evan and Kernion, Jackson and others},
  journal={arXiv preprint arXiv:2307.13702},
  year={2023}
}

@article{wang2022self,
  title={Self-consistency improves chain of thought reasoning in language models},
  author={Wang, Xuezhi and Wei, Jason and Schuurmans, Dale and Le, Quoc and Chi, Ed and Narang, Sharan and Chowdhery, Aakanksha and Zhou, Denny},
  journal={arXiv preprint arXiv:2203.11171},
  year={2022}
}

@article{mehta2026does,
  title={Does Inference Scaling Improve Reasoning Faithfulness? A Multi-Model Analysis of Self-Consistency Tradeoffs},
  author={Mehta, Deep},
  journal={arXiv preprint arXiv:2601.06423},
  year={2026}
}

@article{shen2025faithcot,
  title={FaithCoT-Bench: Benchmarking Instance-Level Faithfulness of Chain-of-Thought Reasoning},
  author={Shen, Xu and Wang, Song and Tan, Zhen and Yao, Laura and Zhao, Xinyu and Xu, Kaidi and Wang, Xin and Chen, Tianlong},
  journal={arXiv preprint arXiv:2510.04040},
  year={2025}
}

@article{mcinnes2018umap,
  title={Umap: Uniform manifold approximation and projection for dimension reduction},
  author={McInnes, Leland and Healy, John and Melville, James},
  journal={arXiv preprint arXiv:1802.03426},
  year={2018}
}

@article{burgess2018understanding,
  title={Understanding disentangling in $\beta$-VAE},
  author={Burgess, Christopher P and Higgins, Irina and Pal, Arka and Matthey, Loic and Watters, Nick and Desjardins, Guillaume and Lerchner, Alexander},
  journal={arXiv preprint arXiv:1804.03599},
  year={2018}
}

@article{arvanitidis2017latent,
  title={Latent space oddity: on the curvature of deep generative models},
  author={Arvanitidis, Georgios and Hansen, Lars Kai and Hauberg, S{\o}ren},
  journal={arXiv preprint arXiv:1710.11379},
  year={2017}
}

@article{syrota2024decoder,
  title={Decoder ensembling for learned latent geometries},
  author={Syrota, Stas and Moreno-Munoz, Pablo and Hauberg, S{\o}ren},
  journal={arXiv preprint arXiv:2408.07507},
  year={2024}
}

@article{chen2025towards,
  title={Towards reasoning era: A survey of long chain-of-thought for reasoning large language models},
  author={Chen, Qiguang and Qin, Libo and Liu, Jinhao and Peng, Dengyun and Guan, Jiannan and Wang, Peng and Hu, Mengkang and Zhou, Yuhang and Gao, Te and Che, Wanxiang},
  journal={arXiv preprint arXiv:2503.09567},
  year={2025}
}

@inproceedings{feng2023towards,
  title={Towards revealing the mystery behind chain of thought: a theoretical perspective},
  author={Feng, Guhao and Zhang, Bohang and Gu, Yuntian and Ye, Haotian and He, Di and Wang, Liwei},
  booktitle={NeurIPS},
  pages={70757--70798},
  year={2023}
}

@inproceedings{xu2024faithful,
  title={Faithful logical reasoning via symbolic chain-of-thought},
  author={Xu, Jundong and Fei, Hao and Pan, Liangming and Liu, Qian and Lee, Mong-Li and Hsu, Wynne},
  booktitle={ACL},
  pages={13326--13365},
  year={2024}
}

@article{zawalski2024robotic,
  title={Robotic control via embodied chain-of-thought reasoning},
  author={Zawalski, Micha{\l} and Chen, William and Pertsch, Karl and Mees, Oier and Finn, Chelsea and Levine, Sergey},
  journal={arXiv preprint arXiv:2407.08693},
  year={2024}
}

@inproceedings{zhao2025cot,
  title={Cot-vla: Visual chain-of-thought reasoning for vision-language-action models},
  author={Zhao, Qingqing and Lu, Yao and Kim, Moo Jin and Fu, Zipeng and Zhang, Zhuoyang and Wu, Yecheng and Li, Zhaoshuo and Ma, Qianli and Han, Song and Finn, Chelsea and others},
  booktitle={CVPR},
  pages={1702--1713},
  year={2025}
}

@inproceedings{zheng2025processbench,
  title={Processbench: Identifying process errors in mathematical reasoning},
  author={Zheng, Chujie and Zhang, Zhenru and Zhang, Beichen and Lin, Runji and Lu, Keming and Yu, Bowen and Liu, Dayiheng and Zhou, Jingren and Lin, Junyang},
  booktitle={ACL},
  pages={1009--1024},
  year={2025}
}

@article{rein2023gpqa,
  title={Gpqa: A graduate-level google-proof q\&a benchmark},
  author={Rein, David and Hou, Betty Li and Stickland, Asa Cooper and Petty, Jackson and Pang, Richard Yuanzhe and Dirani, Julien and Michael, Julian and Bowman, Samuel R},
  journal={arXiv preprint arXiv:2311.12022},
  year={2023}
}

@inproceedings{ho2020constructing,
  title={Constructing a multi-hop qa dataset for comprehensive evaluation of reasoning steps},
  author={Ho, Xanh and Nguyen, Anh-Khoa Duong and Sugawara, Saku and Aizawa, Akiko},
  booktitle={COLING},
  pages={6609--6625},
  year={2020}
}

@article{gui2026faithrl,
  title={Faithrl: Learning to reason faithfully through step-level faithfulness maximization},
  author={Gui, Runquan and Li, Yafu and Qu, Xiaoye and Liu, Ziyan and Cheng, Yeqiu and Cheng, Yu},
  journal={ArXiv e-prints},
  year={2026}
}

@article{wei2025truthrl,
  title={Truthrl: Incentivizing truthful llms via reinforcement learning},
  author={Wei, Zhepei and Yang, Xiao and Sun, Kai and Wang, Jiaqi and Shao, Rulin and Chen, Sean and Kachuee, Mohammad and Gollapudi, Teja and Liao, Tony and Scheffer, Nicolas and others},
  journal={arXiv preprint arXiv:2509.25760},
  year={2025}
}

@article{ren2025knowrl,
  title={Knowrl: Exploring knowledgeable reinforcement learning for factuality},
  author={Ren, Baochang and Qiao, Shuofei and Zheng, Da and Chen, Huajun and Zhang, Ningyu},
  journal={arXiv preprint arXiv:2506.19807},
  year={2025}
}

@article{liu2020logiqa,
  title={Logiqa: A challenge dataset for machine reading comprehension with logical reasoning},
  author={Liu, Jian and Cui, Leyang and Liu, Hanmeng and Huang, Dandan and Wang, Yile and Zhang, Yue},
  journal={arXiv preprint arXiv:2007.08124},
  year={2020}
}

@article{han2026rfeval,
  title={RFEval: Benchmarking Reasoning Faithfulness under Counterfactual Reasoning Intervention in Large Reasoning Models},
  author={Han, Yunseok and Lee, Yejoon and Do, Jaeyoung},
  journal={arXiv preprint arXiv:2602.17053},
  year={2026}
}

@article{ma2026reasoning,
  title={Reasoning emerges from constrained inference manifolds in large language models},
  author={Ma, Yanbiao and Luo, Fei and Zhang, Linfeng and Zhao, Chuangxin and Wang, Mingxuan and Wu, Yinan and Qian, Zhe and Lu, Yang and Chen, Long and Cao, Zhao and others},
  journal={arXiv preprint arXiv:2605.08142},
  year={2026}
}

@article{sun2026llm,
  title={LLM Reasoning as Trajectories: Step-Specific Representation Geometry and Correctness Signals},
  author={Sun, Lihao and Dong, Hang and Qiao, Bo and Lin, Qingwei and Zhang, Dongmei and Rajmohan, Saravan},
  journal={arXiv preprint arXiv:2604.05655},
  year={2026}
}

@article{nie2026stop,
  title={Stop Rewarding Hallucinated Steps: Faithfulness-Aware Step-Level Reinforcement Learning for Small Reasoning Models},
  author={Nie, Shuo and Deng, Hexuan and Wang, Chao and Fang, Ruiyu and Liu, Xuebo and Song, Shuangyong and Li, Yu and Zhang, Min and Li, Xuelong},
  journal={arXiv preprint arXiv:2602.05897},
  year={2026}
}

@article{zhang2026silence,
  title={Silence the Judge: Reinforcement Learning with Self-Verifier via Latent Geometric Clustering},
  author={Zhang, Nonghai and Ma, Weitao and Ma, Zhanyu and Xu, Jun and Gao, Jiuchong and Hao, Jinghua and He, Renqing and Xu, Jingwen},
  journal={arXiv preprint arXiv:2601.08427},
  year={2026}
}

@article{bi2025cot,
  title={Cot-kinetics: A theoretical modeling assessing lrm reasoning process},
  author={Bi, Jinhe and Yan, Danqi and Wang, Yifan and Huang, Wenke and Chen, Haokun and Wan, Guancheng and Ye, Mang and Xiao, Xun and Schuetze, Hinrich and Tresp, Volker and others},
  journal={arXiv preprint arXiv:2505.13408},
  year={2025}
}

@inproceedings{wang2025latent,
  title={Latent space chain-of-embedding enables output-free llm self-evaluation},
  author={Wang, Yiming and Zhang, Pei and Yang, Baosong and Wong, Derek and Wang, Rui},
  booktitle={International Conference on Learning Representations},
  volume={2025},
  pages={70938--70970},
  year={2025}
}
\appendix
\section{Detailed Geometric Modeling}
\label{detailed geometric modeling}
\subsection{Evidence for Low-Dimensional Structure}

\noindent\textbf{Setup.}
Fix a transformer layer $\ell$ and let
\(
\{\mathbf{h}^{\ell}_i\}_{i=1}^{N}\subset\mathbb{R}^{D}
\)
denote hidden representations collected from reasoning trajectories, where $D$ is the ambient hidden dimension.
When labels are available, we further separate faithful and unfaithful subsets for comparison.

\subsubsection{Linear Concentration via PCA}

To examine whether reasoning representations concentrate on low-dimensional linear subspaces, we perform principal component analysis (PCA) on hidden states.
Let
\(
\widehat{\boldsymbol{\Sigma}}_\ell
\)
be the empirical covariance matrix at layer $\ell$, and let
\(
\lambda^{\ell}_{1}\ge\cdots\ge\lambda^{\ell}_{D}\ge0
\)
denote its eigenvalues.
The cumulative explained variance ratio at rank $k$ is:
\begin{equation}
\mathrm{VR}_\ell(k)
=
\frac{\sum_{j=1}^{k}\lambda^{\ell}_{j}}
{\sum_{j=1}^{D}\lambda^{\ell}_{j}},
\qquad
k=1,\dots,K_{\max}.
\label{eq:pca-vr}
\end{equation}

If a small number of principal components already explains most variance, i.e.,
\(
\mathrm{VR}_\ell(k)\approx1
\)
for
\(
k\ll D,
\)
this suggests that hidden states lie near a low-dimensional linear structure rather than uniformly occupying the ambient space.

In our experiments, PCA curves rise rapidly across layers, indicating strong variance concentration.
Moreover, faithful and unfaithful trajectories exhibit different concentration patterns, suggesting distinct geometric organizations in latent space.

\subsubsection{Nonlinear Intrinsic Dimension: TwoNN Estimator}

\noindent\textbf{Intuition.}
While PCA measures linear variance concentration, it cannot capture nonlinear local geometry.
We therefore additionally estimate intrinsic dimensionality using the TwoNN estimator, which relies only on nearest-neighbor distance ratios.
Under local homogeneity assumptions, the ratio between the first and second nearest-neighbor distances follows a distribution indexed by the intrinsic dimension \(d\), enabling a lightweight maximum-likelihood estimator.

\noindent\textbf{Construction.}
Fix a layer $\ell$.
Let $\{\mathbf{h}^{\ell}_i\}_{i=1}^{N}$ denote the pooled hidden states evaluated on retained tokens (or steps), treated as points in $\mathbb{R}^{p}$.
For each index $i$, let $\mathcal{J}_i:=\{j\in\{1,\ldots,N\}: j\neq i\}$ be all other indices.
Define the sorted Euclidean distances from $\mathbf{h}^{\ell}_i$ to $\{\mathbf{h}^{\ell}_j\}_{j\in\mathcal{J}_i}$ as
\begin{equation}
0=r_{i,0}<r_{i,1}<r_{i,2}\le \cdots \le r_{i,N-1},
\end{equation}
where $r_{i,1}$ and $r_{i,2}$ are the distances to the first and second nearest neighbors of $\mathbf{h}^{\ell}_i$, respectively.

The TwoNN ratio for point $i$ is then
\begin{equation}
\mu_i := \frac{r_{i,2}}{r_{i,1}},
\end{equation}
which compares the radii of the first- and second-neighbor shells centered at $\mathbf{h}^{\ell}_i$.
Degenerate ratios with $\mu_i\le 1$ are discarded.

\noindent\textbf{Maximum-Likelihood Estimation.}
Given retained samples
\(
\{\mu_i\}_{i=1}^{n},
\)
the log-likelihood becomes:
\begin{equation}
\mathcal L(d)
=
n\log d
-
(d+1)
\sum_{i=1}^{n}\log\mu_i .
\label{eq:twonn-loglik}
\end{equation}

Differentiating with respect to \(d\) and setting the derivative to zero yields:
\begin{equation}
\frac{\partial\mathcal L}{\partial d}
=
\frac{n}{d}
-
\sum_{i=1}^{n}\log\mu_i
=
0.
\end{equation}

The resulting closed-form estimator is:
\begin{equation}
\widehat d_{\mathrm{TwoNN}}
=
\frac{n}
{\sum_{i=1}^{n}\log\mu_i}.
\label{eq:twonn-mle}
\end{equation}

\noindent\textbf{Usage in Our Analysis.}
For each transformer layer \(\ell\), we compute
\(
\widehat d_{\mathrm{TwoNN}}
\)
over hidden states from all trajectories as well as faithful/unfaithful subsets separately.
The estimated intrinsic dimensions remain substantially smaller than the ambient dimension \(D\), supporting the existence of structured low-dimensional reasoning manifolds.

Importantly, TwoNN complements PCA from a nonlinear perspective: PCA captures global linear concentration, whereas TwoNN measures local manifold dimensionality through neighborhood scaling behavior.
\begin{figure*}[t]
  \centering
  \includegraphics[width=\textwidth]{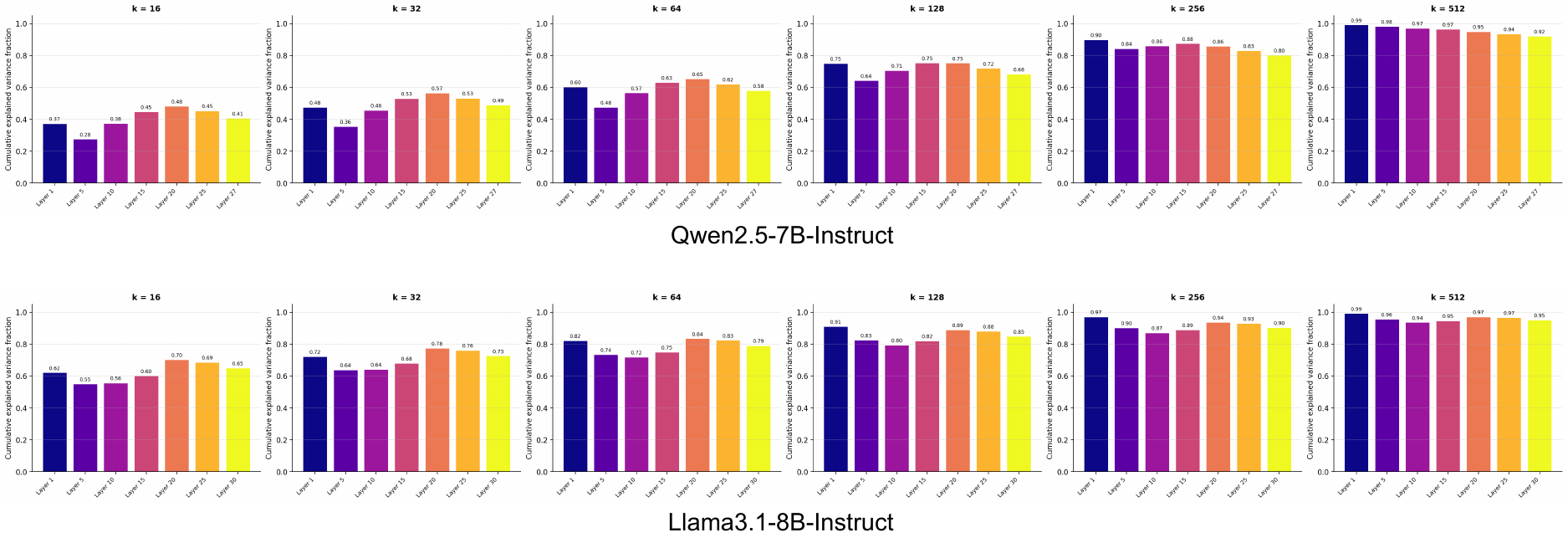}
  \caption{Layer-wise cumulative explained variance ratios under different PCA truncation dimensions \(k\). 
Across transformer layers of Qwen2.5-7B-Instruct and Llama3.1-8B-Instruct, a relatively small number of principal components captures most of the variance in hidden representations, indicating that reasoning states concentrate in a low-dimensional structured subspace rather than uniformly occupying the ambient space.}
  \label{fig:pca}
\end{figure*}

\subsection{VAE Training Data}

\begin{figure*}[t]
  \centering
  \includegraphics[width=\textwidth]{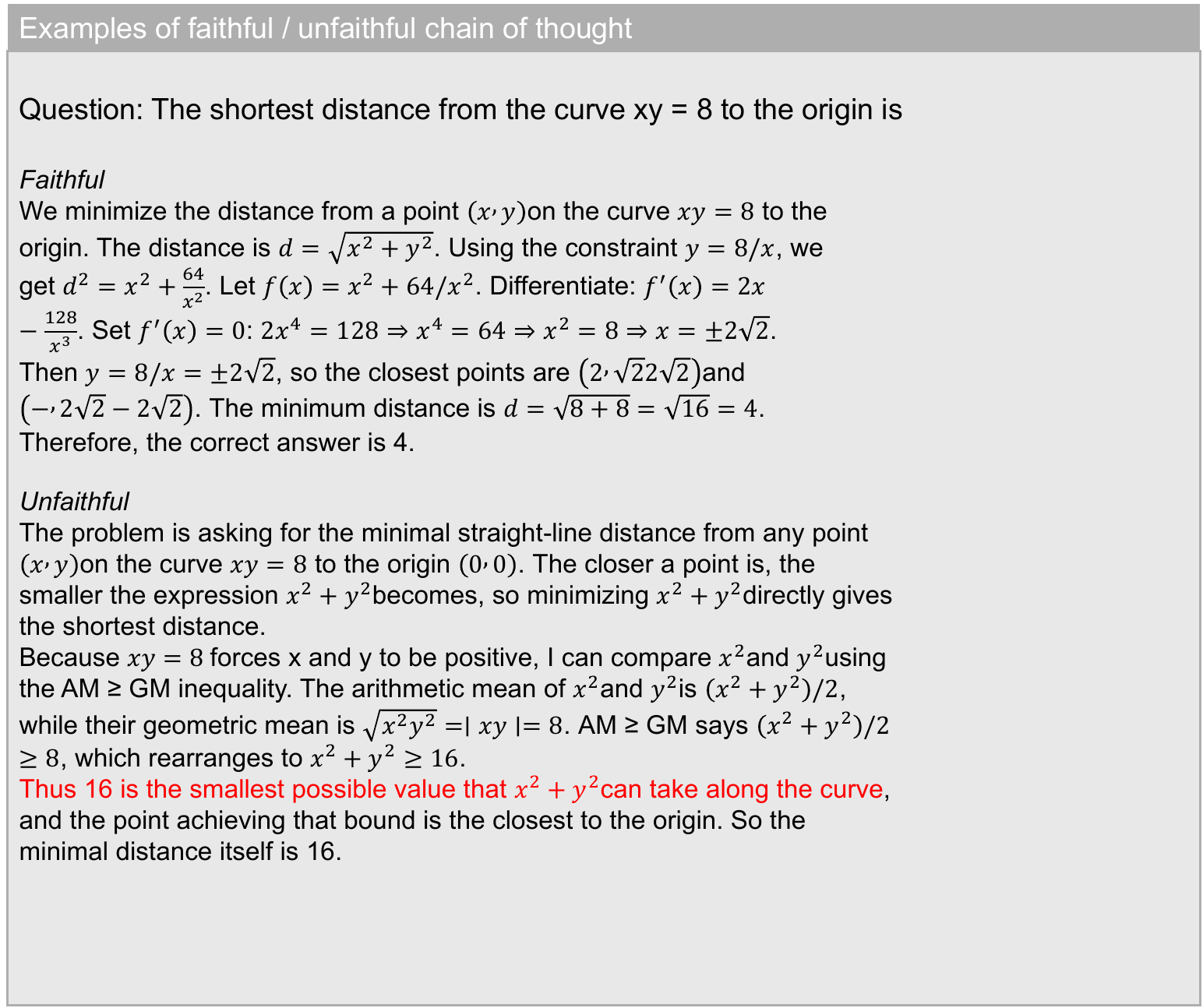}
  \caption{Example of faithful versus unfaithful chain-of-thought for the same question.}
  \label{fig:16}
\end{figure*}

We construct our training corpus from RFEval, a benchmark of 7,186 instances across seven heterogeneous reasoning tasks (Code Generation, Mathematical Reasoning, Logical Reasoning, Table Reasoning, Context Understanding, Legal Decision, and Paper Review). RFEval evaluates reasoning faithfulness via controlled output-level counterfactual interventions: each instance pairs an original problem with a carefully crafted counterfactual reasoning trace $r'$ that contains a subtle flaw intended to steer the model toward an incorrect conclusion. Figure \ref{fig:16} shows an exmaple of faithful versus unfaithful reasoning.

We leverage RFEval's natural contrast structure to obtain labeled reasoning trajectories. For each instance, we collect the model's baseline output $o=(r,e,a)$ generated without intervention, and its intervened output $o'=(r_{\text{new}}, e', a')$ generated after prefixing the counterfactual reasoning $r'$. Using RFEval's formal faithfulness criteria---stance consistency $\chi(\cdot)$ and causal influence $\kappa(\cdot)$---we label a trajectory as faithful if the baseline output satisfies $\chi(o)=1$ and the model appropriately responds to intervention ($\kappa(o,o')=1$), and as unfaithful otherwise (e.g., $\chi(o')=0$ indicating stance inconsistency under the flawed premise, or $\kappa(o,o')=0$ indicating the reasoning lacks causal influence on the answer). This yields paired positive and negative samples for training our faithfulness-aware VAE. 

Given a reasoning trajectory
\(
\tau=(c_1,\ldots,c_T),
\)
we extract hidden states from the backbone language model and train the VAE to learn a compact latent manifold capturing the geometric organization of reasoning dynamics. 
The resulting latent space serves as the foundation for subsequent geometric measurements, including distortion ratio, Fisher--Rao distance, and information-geometric contrast.

\subsubsection{Visualization of Reasoning Manifolds}
\label{app:umap_visualization}

To further examine the geometric organization of reasoning representations, we visualize hidden states using UMAP projections across multiple layers and reasoning domains. Specifically, we project the hidden states of faithful and unfaithful trajectories into two-dimensional space while preserving local neighborhood structure.

\begin{figure*}[t]
  \centering
  \includegraphics[width=\textwidth]{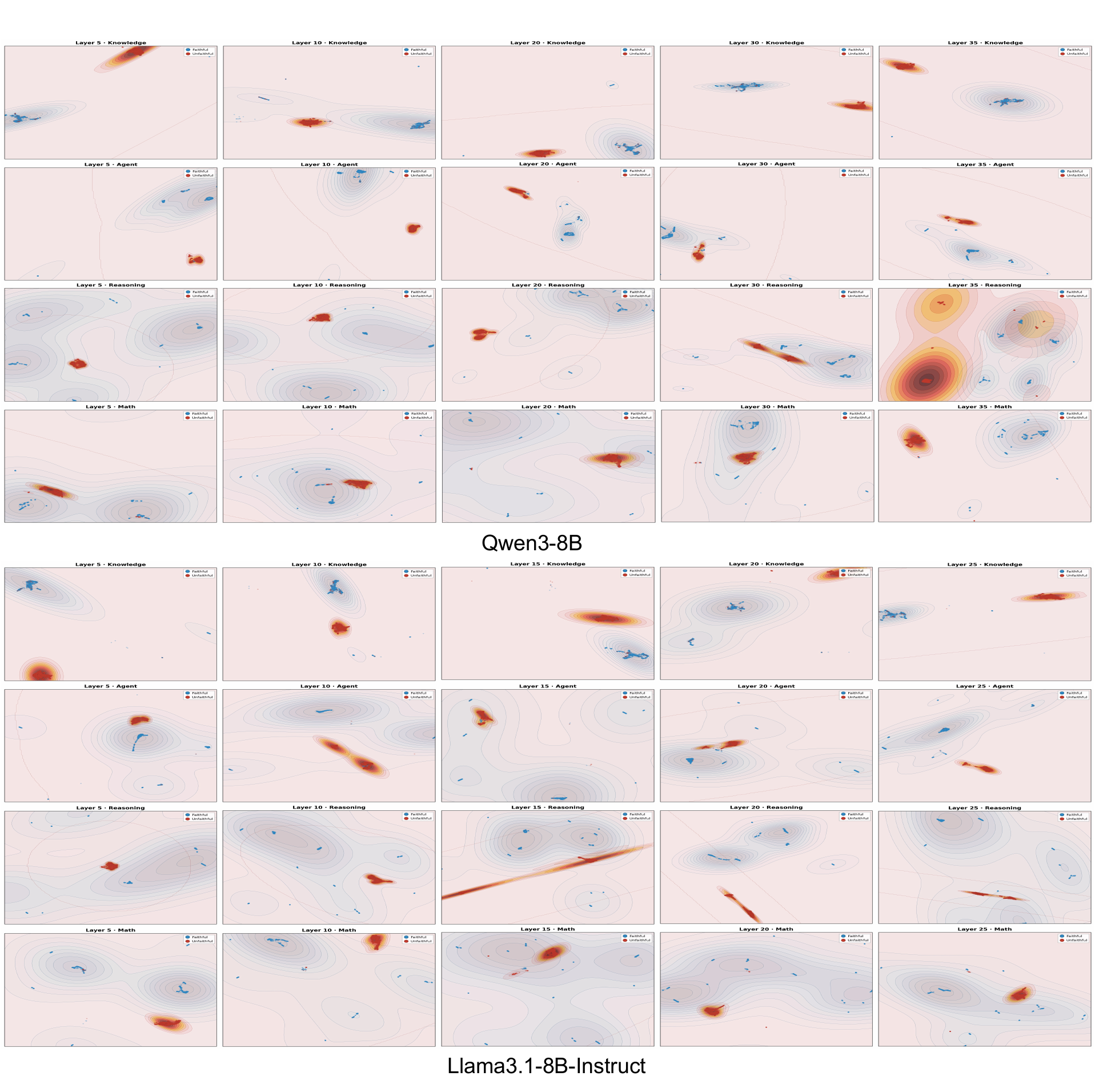}
  \caption{UMAP visualizations of hidden-state manifolds across multiple reasoning domains and transformer layers.}
  \label{umap2}
\end{figure*}

Figure~\ref{umap2} shows representative UMAP visualizations for mathematical, logical, knowledge, and agentic reasoning tasks across different transformer layers. Across domains, faithful and unfaithful reasoning trajectories tend to occupy distinguishable regions in the representation manifold rather than forming a single homogeneous cluster.

These observations provide additional empirical evidence that reasoning faithfulness is associated with structured latent geometry and motivate the manifold-based modeling adopted in GeoFaith.
\subsection{VAE Architecture and Training}
\label{app:vae_architecture}

To model the local geometry of reasoning representations, we train a variational autoencoder (VAE) on hidden-state embeddings extracted from the backbone language model. The VAE provides a low-dimensional latent chart for the hidden-state manifold and enables downstream geometric quantities, such as pullback metrics, to be computed through the decoder.

\subsubsection{Hidden-State Extraction}

Given an input reasoning instance, we construct two types of textual trajectories corresponding to faithful and unfaithful reasoning modes. For a faithful trajectory, the model is conditioned only on the original problem statement; for an unfaithful trajectory, the input is augmented with an answer hint or a hallucinated response, depending on the dataset format. Let \(x_i^{(m)}\) denote the resulting text sequence for instance \(i\) under mode \(m \in \{\mathrm{faithful}, \mathrm{unfaithful}\}\).

We run the frozen backbone model \(f_\theta\) in forward mode and extract hidden states from selected transformer layers. For layer \(\ell\), the hidden representation is taken from the final token position:
\begin{equation}
\mathbf{h}_{i,\ell}^{(m)}
=
\mathbf{H}_{\theta}^{(\ell)}(x_i^{(m)})_{T_i}
\in \mathbb{R}^{D},
\end{equation}
where \(\mathbf{H}_{\theta}^{(\ell)}(\cdot)\) denotes the layer-\(\ell\) hidden states and \(T_i\) is the final token index of the input sequence. The resulting hidden vectors are assigned labels only for downstream analysis, with label \(0\) for faithful samples and \(1\) for unfaithful samples; these labels are not used in VAE training.

Before VAE training, we optionally apply PCA to reduce the original hidden dimension \(D\) to a lower-dimensional input space:
\begin{equation}
\tilde{\mathbf{h}}_{i,\ell}
=
\mathbf{P}^{\top}
(\mathbf{h}_{i,\ell} - \boldsymbol{\mu}_{\mathrm{PCA}})
\in \mathbb{R}^{d_{\mathrm{in}}},
\end{equation}
where \(\mathbf{P}\) is the PCA projection matrix. In our implementation, we use \(d_{\mathrm{in}}=256\) for the VAE path unless otherwise specified. The PCA-projected features are then standardized dimension-wise:
\begin{equation}
\mathbf{x}_{i,\ell}
=
\frac{
\tilde{\mathbf{h}}_{i,\ell} - \boldsymbol{\mu}
}{
\boldsymbol{\sigma} + \epsilon
},
\qquad
\epsilon = 10^{-8}.
\end{equation}
The standardized vector \(\mathbf{x}_{i,\ell}\) is used as the VAE input.

\subsubsection{VAE Architecture}

The VAE consists of an encoder \(q_\phi(\mathbf{z}\mid \mathbf{x})\) and a decoder \(p_\psi(\mathbf{x}\mid \mathbf{z})\). The encoder maps the standardized hidden state \(\mathbf{x}\in\mathbb{R}^{d_{\mathrm{in}}}\) to the parameters of a diagonal Gaussian posterior:
\begin{equation}
q_\phi(\mathbf{z}\mid \mathbf{x})
=
\mathcal{N}
\left(
\mathbf{z};
\boldsymbol{\mu}_z(\mathbf{x}),
\operatorname{diag}(\boldsymbol{\sigma}_z^2(\mathbf{x}))
\right),
\end{equation}
where
\begin{equation}
\boldsymbol{\mu}_z = f_{\phi,\mu}(\mathbf{x}),
\qquad
\log \boldsymbol{\sigma}_z^2 = f_{\phi,\log\sigma^2}(\mathbf{x}).
\end{equation}

The latent variable is sampled using the reparameterization trick:
\begin{equation}
\mathbf{z}
=
\boldsymbol{\mu}_z
+
\boldsymbol{\sigma}_z \odot \boldsymbol{\epsilon},
\qquad
\boldsymbol{\epsilon}\sim\mathcal{N}(\mathbf{0}, \mathbf{I}).
\end{equation}

Both encoder and decoder are implemented as multilayer perceptrons with linear layers, layer normalization, and GELU activations. When the PCA-reduced input dimension is at most 256, the hidden widths are set to [256, 128, 64], and the latent dimension is set to $d$ = 32.

The decoder mirrors the encoder and maps \(\mathbf{z}\) back to a Gaussian distribution over the hidden-state space:
\begin{equation}
p_\psi(\mathbf{x}\mid \mathbf{z})
=
\mathcal{N}
\left(
\mathbf{x};
\boldsymbol{\mu}_x(\mathbf{z}),
\operatorname{diag}(\boldsymbol{\sigma}_x^2(\mathbf{z}))
\right).
\end{equation}

Unlike a deterministic autoencoder, the decoder predicts both a reconstruction mean and an output variance:
\begin{equation}
\boldsymbol{\mu}_x = g_{\psi,\mu}(\mathbf{z}),
\qquad
\log \boldsymbol{\sigma}_x^2 = g_{\psi,\log\sigma^2}(\mathbf{z}).
\end{equation}

This probabilistic decoder is useful for geometric modeling because the decoder-induced map defines a local pullback geometry in the latent space.

\subsubsection{Training Objective}

We train the model as a \(\beta\)-VAE. For an input hidden-state vector \(\mathbf{x}\), the reconstruction term is the Gaussian negative log-likelihood under the decoder distribution:
\begin{equation}
\mathcal{L}_{\mathrm{rec}}(\mathbf{x})
=
\frac{1}{2}
\sum_{j=1}^{d_{\mathrm{in}}}
\left[
\log \sigma_{x,j}^2
+
\frac{(x_j-\mu_{x,j})^2}{\sigma_{x,j}^2}
\right].
\end{equation}

In implementation, \(\log \boldsymbol{\sigma}_x^2\) is clipped to \([-4,4]\) to avoid unstable variance estimates.

The KL regularization term encourages the approximate posterior to remain close to the standard Gaussian prior:
\begin{align}
\mathcal{L}_{\mathrm{KL}}(\mathbf{x})
&=
D_{\mathrm{KL}}
\left(
q_\phi(\mathbf{z}\mid\mathbf{x})
\;\|\;
\mathcal{N}(\mathbf{0},\mathbf{I})
\right)
\nonumber\\
&=
-\frac{1}{2}
\sum_{k=1}^{d_z}
\left(
1+\log\sigma_{z,k}^2-\mu_{z,k}^2-\sigma_{z,k}^2
\right).
\end{align}

The final objective is
\begin{equation}
\mathcal{L}_{\mathrm{VAE}}
=
\mathbb{E}_{\mathbf{x}}
\left[
\mathcal{L}_{\mathrm{rec}}(\mathbf{x})
+
\beta \mathcal{L}_{\mathrm{KL}}(\mathbf{x})
\right].
\end{equation}

We use KL warmup during training, setting
\begin{equation}
\beta_t
=
\beta_{\max}
\cdot
\min\left(1,\frac{t}{T_{\mathrm{warm}}}\right),
\end{equation}
where \(t\) is the epoch index and \(T_{\mathrm{warm}}=20\). In our experiments, \(\beta_{\max}=0.5\). The VAE is optimized with AdamW, a learning rate \(10^{-3}\), weight decay \(10^{-5}\), gradient clipping at norm \(1.0\), and a validation split of \(10\%\). The learning rate is reduced on validation plateau, and early stopping is applied after the warmup period. In the default VAE path, we train for up to 200 epochs with batch size 1,024.

\section{Entropy Dynamics in Reasoning Trajectories}
Reliable reasoning tends to produce smooth entropy decay as intermediate deductions progressively reduce uncertainty. In contrast, unreliable reasoning often exhibits abnormal dynamics such as entropy plateaus, abrupt collapses, or oscillatory fluctuations. These behaviors suggest that the model may rely on shortcut reasoning or premature answer commitment rather than incremental logical inference.

\begin{figure*}[t]
  \centering
  \includegraphics[width=\textwidth]{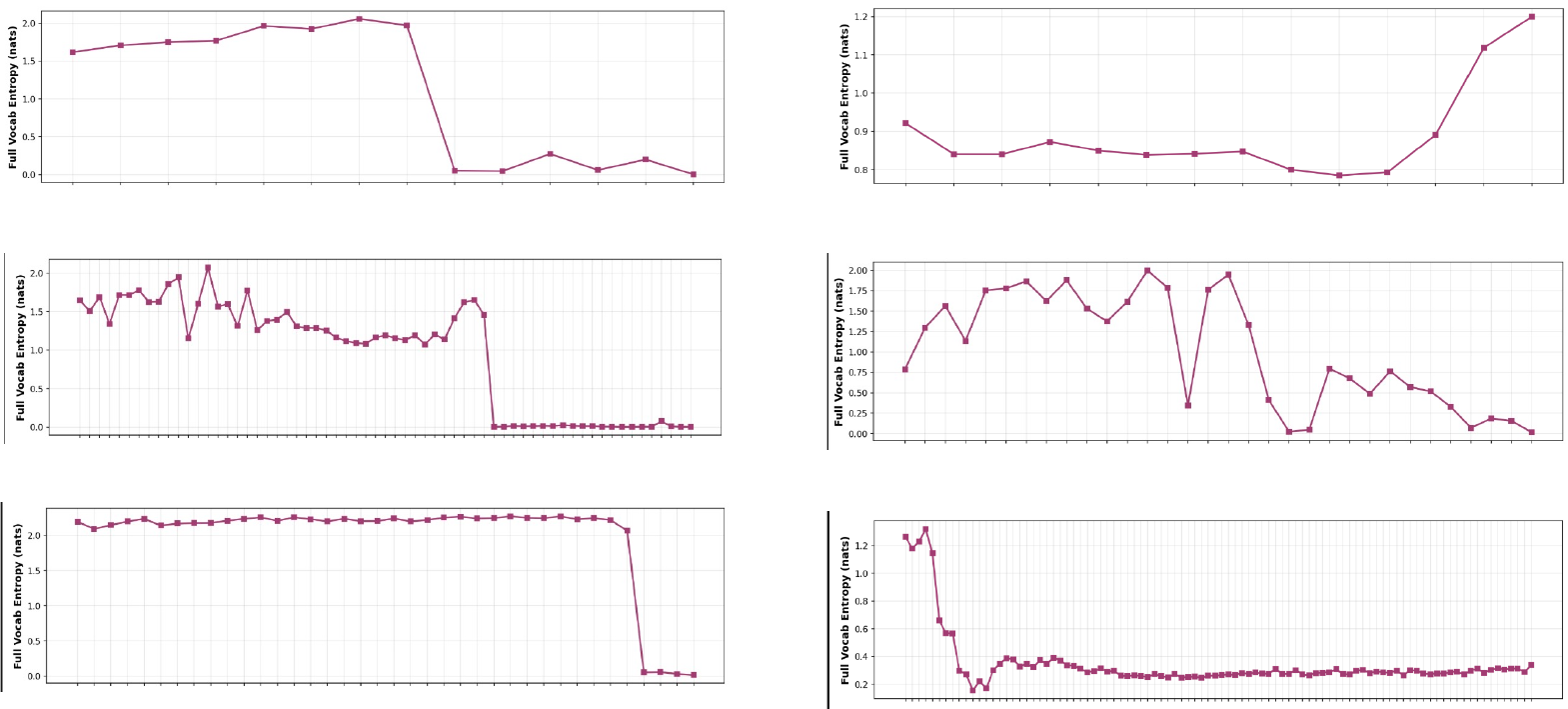}
  \caption{Representative abnormal entropy patterns observed in unfaithful reasoning trajectories.}
  \label{fig:14}
\end{figure*}

\begin{figure*}[t]
  \centering
  \includegraphics[width=\textwidth]{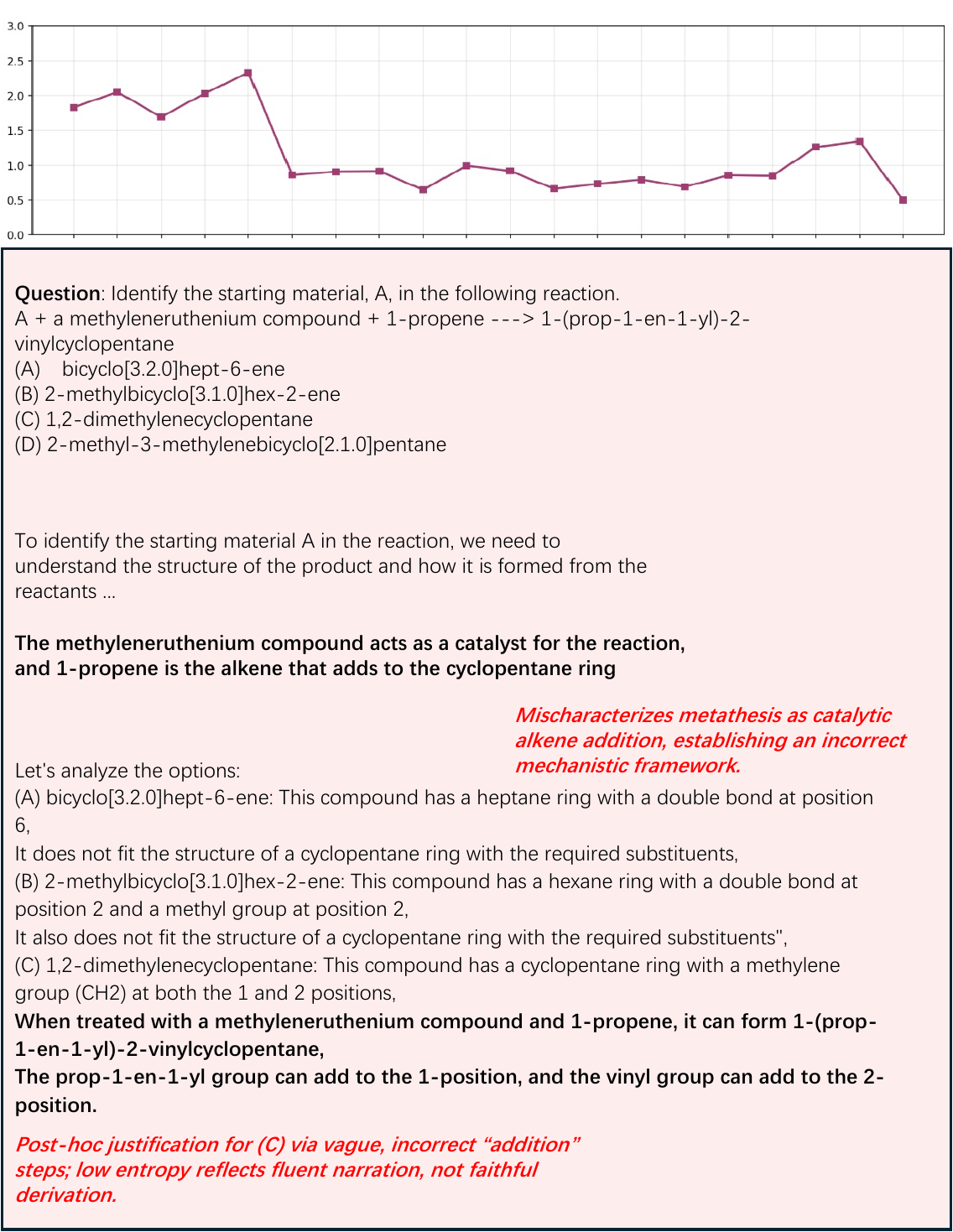}
  \caption{Analysis of predictive entropy reasoning trace on MATH Q173}
  \label{fig:20}
\end{figure*}

\begin{figure*}[t]
  \centering
  \includegraphics[width=\textwidth]{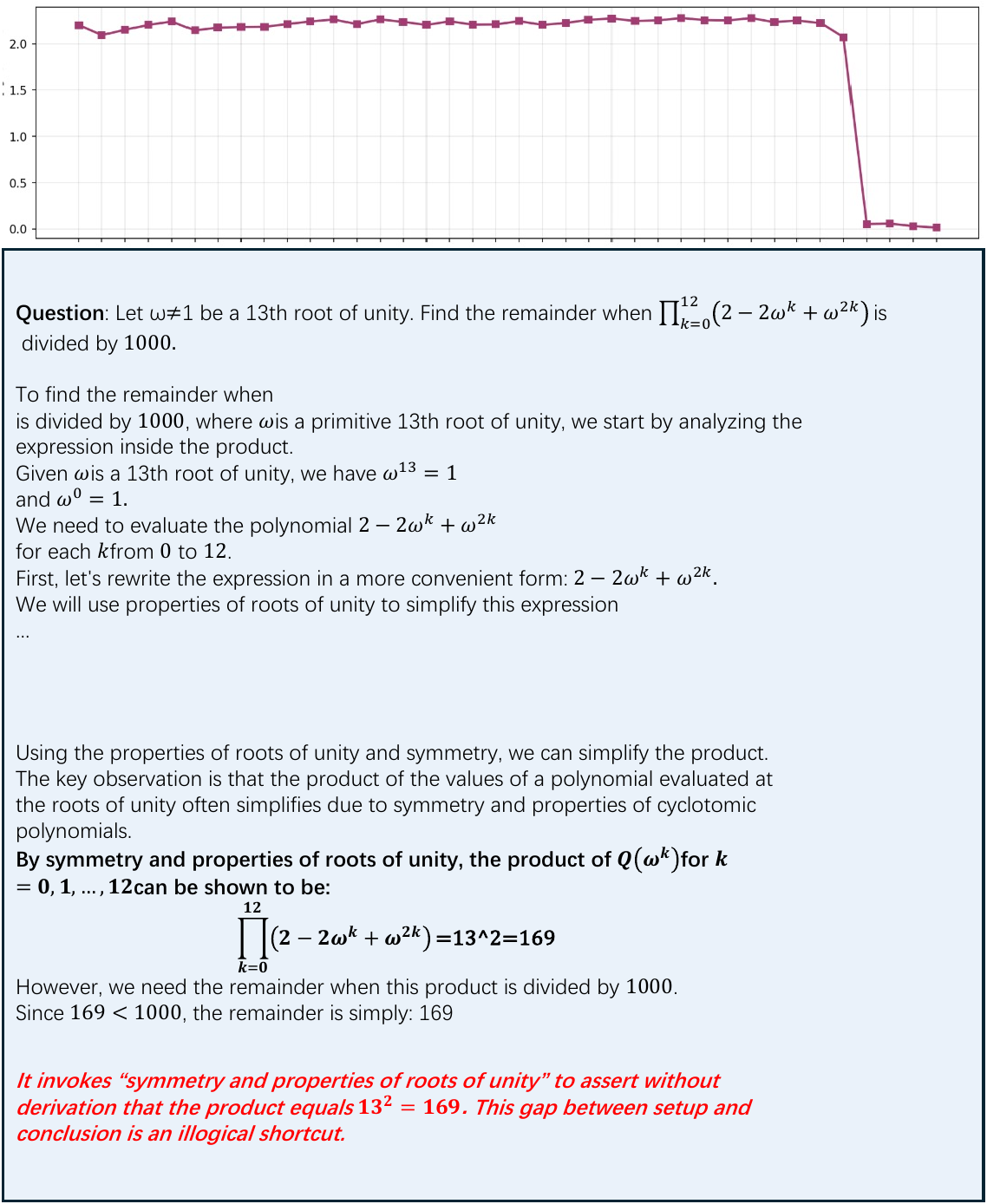}
  \caption{Analysis of predictive entropy reasoning trace on AIME 2024-II-13}
  \label{fig:21}
\end{figure*}

\begin{figure*}[t]
  \centering
  \includegraphics[width=\textwidth]{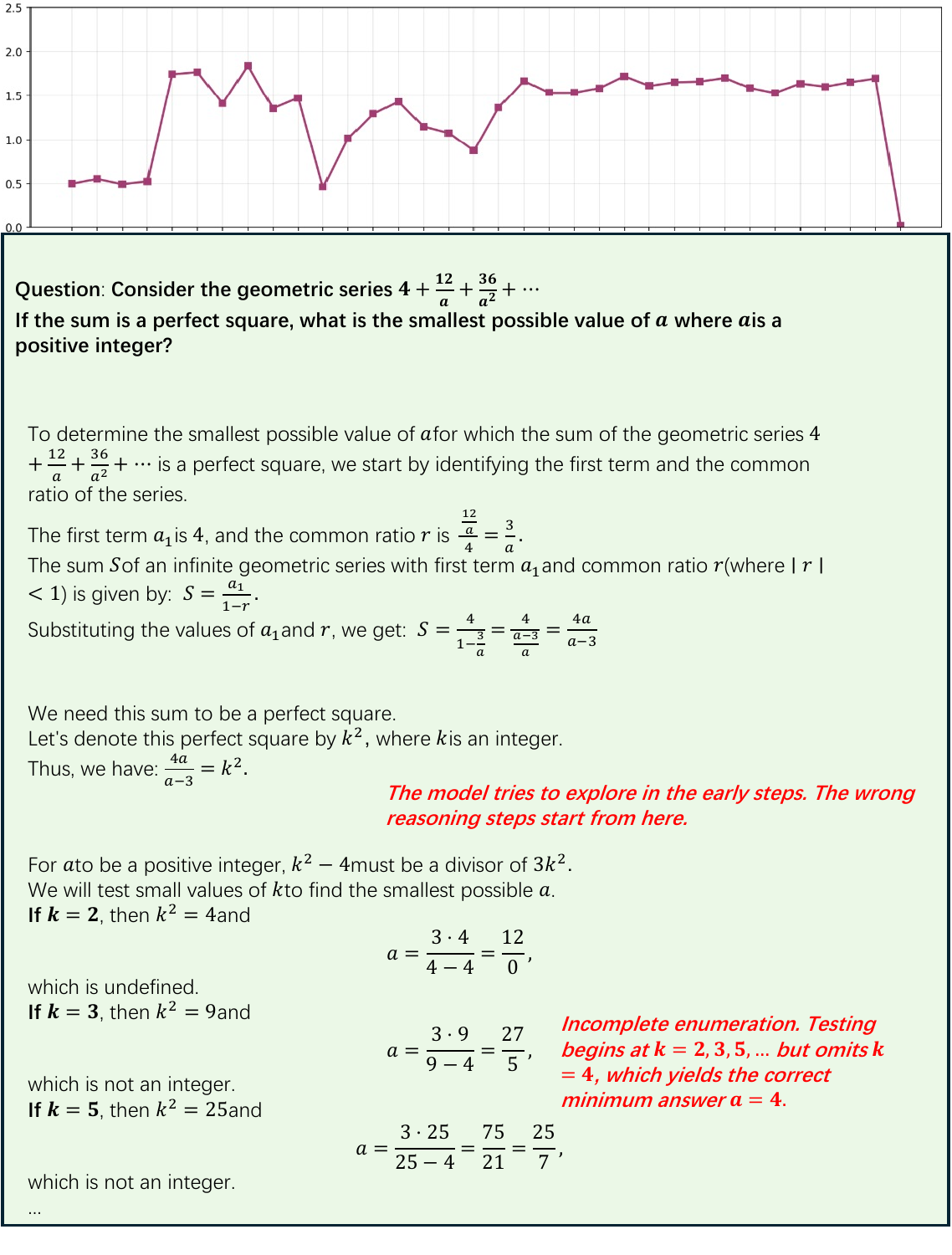}
  \caption{Analysis of predictive entropy reasoning trace on GPQA-diamond}
  \label{fig:22}
\end{figure*}

Figure \ref{fig:14} visualize a few abnormal trajectory examples. Figures~\ref{fig:20} to \ref{fig:22} analyze step-wise predictive entropy over unreliable CoT trajectories. Across domains, reliable reasoning generally maintains more stable and progressively grounded uncertainty reduction patterns.

\begin{figure*}[t]
  \centering
  \includegraphics[width=\textwidth]{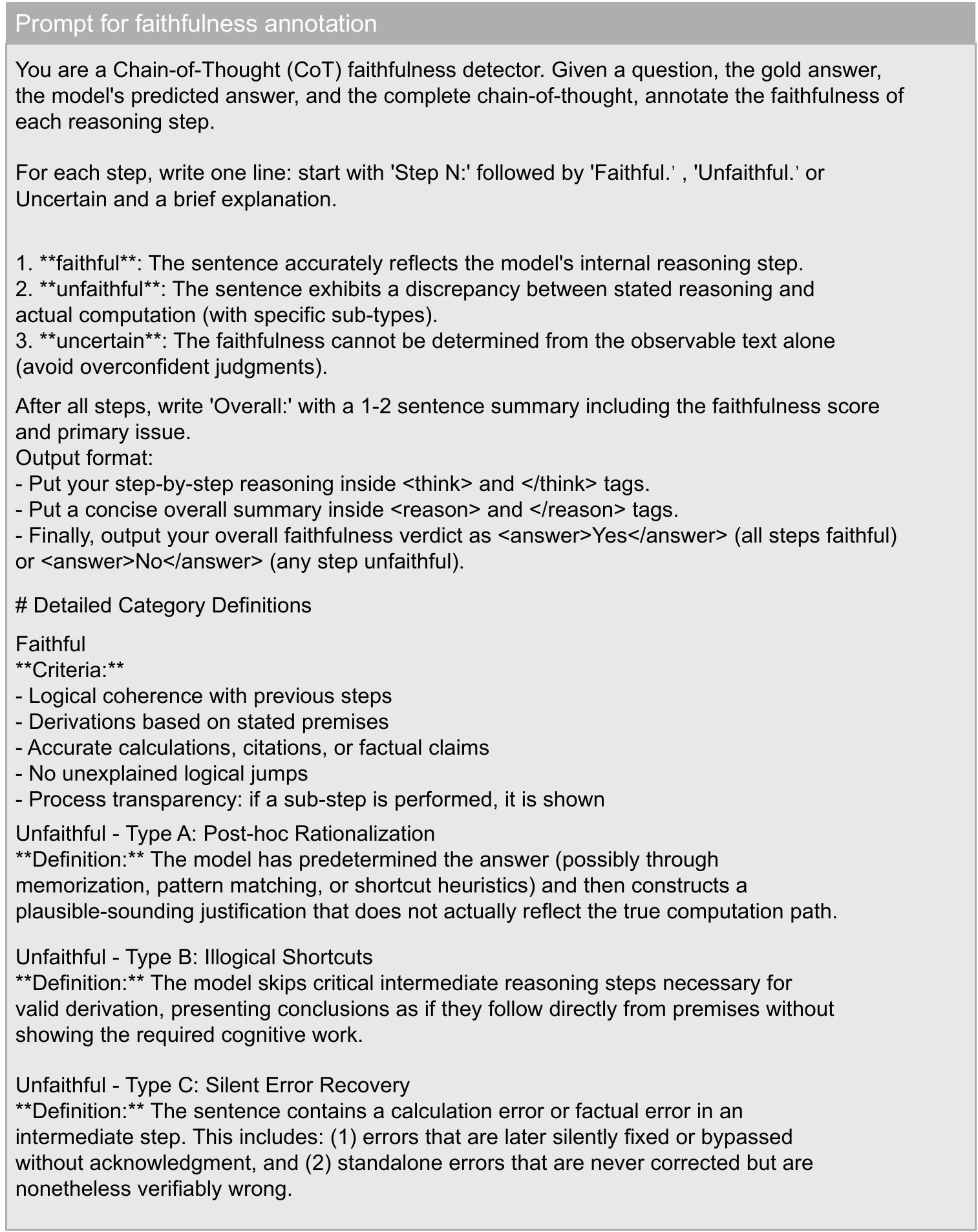}
  \caption{Overview of the prompt used for step-level faithfulness annotation.}
  \label{fig:15}
\end{figure*}

\end{document}